\documentclass[11pt]{article}
\usepackage[hyperref]{ccl2022-en}
\usepackage{times}
\usepackage{url}
\usepackage{latexsym}
\usepackage{fancyhdr}

\title{Through the Lens of Core Competency: Survey on Evaluation of Large Language Models}

\author{Ziyu Zhuang, Qiguang Chen, Longxuan Ma, \textbf{Mingda Li}, \textbf{Yi Han}, \textbf{Yushan Qian},\\
\textbf{Haopeng Bai}, \textbf{Zixian Feng}, \textbf{Weinan Zhang\thanks{*Corresponding author}, Ting Liu}\\
         Research Center for Social Computing and Information Retrieval, \\
         Harbin Institute of Technology\\
         \{zyzhuang, qgchen, lxma, mdli, yihan, ysqian, hpbai, zxfeng, wnzhang, tliu\}@ir.hit.edu.cn}

\date{}

\begin{document}
\maketitle
\begin{abstract}
  From pre-trained language model (PLM) to large language model (LLM), the field of natural language processing (NLP) has witnessed steep performance gains and wide practical uses. The evaluation of a research field guides its direction of improvement. However, LLMs are extremely hard to thoroughly evaluate for two reasons. First of all, traditional NLP tasks become inadequate due to the excellent performance of LLM. Secondly, existing evaluation tasks are difficult to keep up with the wide range of applications in real-world scenarios. To tackle these problems, existing works proposed various benchmarks to better evaluate LLMs. To clarify the numerous evaluation tasks in both academia and industry, we investigate multiple papers concerning LLM evaluations. We summarize 4 core competencies of LLM, including reasoning, knowledge, reliability, and safety. For every competency, we introduce its definition, corresponding benchmarks, and metrics. Under this competency architecture, similar tasks are combined to reflect corresponding ability, while new tasks can also be easily added into the system. Finally, we give our suggestions on the future direction of LLM's evaluation.
\end{abstract}

\section{Introduction}
\label{intro}

\cclfootnote{
    %
    % for review submission
    %
    \hspace{-0.65cm}  % space normally used by the marker
    % Place licence statement here for the camera-ready version. See Section~\ref{licence} of the instructions for preparing a manuscript.
    \textcopyright 2023 China National Conference on Computational Linguistics

    \noindent Published under Creative Commons Attribution 4.0 International License
}

Large language models(LLMs) have achieved great progresses in many areas. One representative, ChatGPT\footnote{https://openai.com/blog/chatgpt/}, which applies the ability of LLMs in the form of dialogue, has received much attention due to its incredible versatility such as creative writing, coding, planning, etc. The evaluation of such a model thus becomes necessary to benchmark and build up its ability while preventing potential harmfulness. 

Existing works on the evaluation of LLMs can be divided into three paradigms. The first line of work is evaluating LLMs with traditional NLP tasks like dialogue, summarization, etc. Since LLMs are actually pre-trained language models(PLMs) with huge model parameter size and data size \cite{DBLP:journals/corr/abs-2001-08361},  benchmarks like GLUE \cite{DBLP:conf/iclr/WangSMHLB19}, SuperGLUE \cite{DBLP:conf/nips/WangPNSMHLB19} can be adopted to evaluate its language understanding ability. The problem is that LLMs work really well on less restrictive tasks like translation, summarization, and natural language understanding tasks. Sometimes LLMs generated outputs' third-party scores are even higher than human generations \cite{DBLP:journals/corr/abs-2211-09110}, showing the need for higher-quality tasks.
Secondly, advanced ability evaluations are proposed to completely test language models. The parameter size difference between LLMs and PLMs brings an amazing phenomenon, emergence \cite{DBLP:journals/tmlr/WeiTBRZBYBZMCHVLDF22,DBLP:journals/corr/abs-2206-04615}, which means that scaled models exhibit abilities that are not possessed in small-scaled language models. For instance, in tasks like reasoning, and tool manipulation, the correlation curve between the number of model parameters and the task effect is non-linear. And the effect will rise sharply when the model parameter exceeds a certain parameter scale. They're called "advanced" because they're more closely related to human abilities and harder for models to complete \cite{DBLP:journals/corr/abs-2304-06364}.
Thirdly, test language models' intrinsic abilities independent of the specific tasks. It can be tested in parallel with almost every task above. Robustness is a classic ability in this paradigm. Due to the black-box nature of neural networks \cite{DBLP:journals/corr/SzegedyZSBEGF13}, robustness problems exist for every modality of input data(vision, audio, test, etc.). 

Current evaluation benchmarks \cite{DBLP:journals/corr/abs-2211-09110,DBLP:journals/corr/abs-2206-04615,eval-harness,DBLP:journals/corr/abs-2304-06364,DBLP:journals/corr/abs-2306-09212} are mostly a mixture of the former three paradigms. They emphasize a complete system of evaluation tasks, in which all tasks are of equal importance. But the significance of marginal increases in model effects on tasks with excellent performance is debatable. Thus numerous evaluation tasks and benchmarks are proposed to follow and challenge the ever-evolving LLMs, while, oddly, seldom being reviewed in a systematic way. How to link numerous tasks and benchmarks, better present the evaluation results, and thus facilitate the research of LLMs is an urgent problem. 

An ideal large language model needs to be capable, reliable, and safe \cite{DBLP:conf/nips/Ouyang0JAWMZASR22}. One surely needs extensive tests on multiple datasets to meet these miscellaneous standards. Moreover, to avoid the prevalent training set leakage, test sets also should be updated regularly \cite{DBLP:journals/corr/abs-2305-08322}. This is similar to the competency \cite{hoffmann1999meanings} tests adopted in corporate recruitment. In competency tests, different task sets are combined to test the corresponding competency. And task sets also need renewal to prevent possible fraud. 

In this survey, \textbf{we draw on the concept of the core competency to integrate multiple evaluation research for LLMs.} We investigated \textbf{540+} tasks widely used in various papers, aggregating tasks corresponding to a certain competency. During this process, 4 core competencies are summarized, including knowledge, reasoning, reliability, and safety. We will introduce the definition, taxonomy, and metrics for these competencies. Through this competency test, superabundant evaluation tasks and benchmarks are combed and clarified for their aiming utility. Furthermore, the evaluation results presented with this procedure will be direct, concise, and focused. Updated new tasks can also be added comprehensively. To support the community in taking this competency test further, We also create an extensible project, which will show the many-to-many relationship between competencies and tasks precisely\footnote{https://github.com/HITSCIR-DT-Code/Core-Competency-Test-for-the-Evaluation-of-LLMs}. Due to the length of the paper, we can only present part of the surveyed results in this paper. A more comprehensive study will be released in a later version.% Notice that the datasets are usually related to multiple evaluation dimensions, we only show the evaluation dimension that the dataset is most concerned about when proposed. 

%
% The following footnote without marker is needed for the camera-ready
% version of the paper.
% Comment out the instructions (first text) and uncomment the 8 lines
% under "final paper" for your variant of English.
%

\section{Core Competencies}
In this section, we introduce the definition and taxonomy of the core competencies we summarized.

\subsection{Knowledge}
%\subsection{Knowledge}
\label{knowledge}
Knowledge is generally defined as the cognition of humans when practicing in the subjective and objective world, which is verified and can be reused over time\footnote{https://plato.stanford.edu/entries/epistemology/}. The large language models (LLMs) nowadays obtain human knowledge from a large scale of training corpus, so that it can use the knowledge to solve various downstream tasks. In this section, we focus on the fundamental knowledge competency of LLMs that facilitates communication and other downstream tasks (such as reasoning). Specifically, we divide the fundamental knowledge into \textbf{linguistic knowledge} and \textbf{world knowledge} \cite{day1998extensive} and introduce the definitions of them and the benchmarks that can evaluate them.

\subsubsection{Linguistic Knowledge Competency}
Linguistic knowledge includes grammatical, semantic, and pragmatic knowledge \cite{fromkin2018introduction}. The grammar of a natural language is its set of structural constraints on speakers' or writers' composition of clauses, phrases, and words. The term can also refer to the study of such constraints, a field that includes domains such as phonology, morphology, and syntax, often complemented by phonetics, semantics, and pragmatics. Semantic \cite{austin1975things} studies the meaning of words, phrases, and sentences, focusing on general meanings rather than on what an individual speaker may want them to mean. Pragmatics \cite{austin1975things} studies language use and how listeners bridge the gap between sentence meaning and the speaker’s meaning. It is concerned with the relationship between semantic meaning, the context of use, and the speaker’s meaning. 

The Linguistic Knowledge competency is embodied in almost all NLP tasks, researchers usually design specific scenarios to test the linguistic competency of LLMs. Some examples are shown in the upper group of Table \ref{knowledge-datasets}. BLiMP \cite{DBLP:journals/tacl/WarstadtPLMPWB20} evaluates what language models (LMs) know about major grammatical phenomena. Linguistic\_mappings \footnote{https://github.com/google/BIG-bench/blob/main/bigbench/benchmark\_tasks/linguistic\_mappings} task aims to explore the depth of linguistic knowledge in enormous language models trained on word prediction. It aims to discover whether such knowledge is structured so as to support the use of grammatical abstractions, both morphological (past tense formation and pluralization) and syntactic (question formation, negation, and pronominalization). The minute\_mysteries\_qa \footnote{https://github.com/google/BIG-bench/blob/main/bigbench/benchmark\_tasks/minute\_mysteries\_qa} is a reading comprehension task focusing on short crime and mystery stories where the goal is to identify the perpetrator and to explain the reasoning behind the deduction and the clues that support it. The metaphor\_boolean \footnote{https://github.com/google/BIG-bench/tree/main/bigbench/benchmark\_tasks/metaphor\_boolean} task presents a model with a metaphoric sentence and asks it to identify whether a second sentence is the correct interpretation of the first. The last three are selected from BIG-Bench \cite{DBLP:journals/corr/abs-2206-04615}, containing diverse task topics including linguistics.

\subsubsection{World Knowledge Competency}
World knowledge is non-linguistic information that helps a reader or listener interpret the meanings of words and sentences \cite{DBLP:books/daglib/0028671}. It is also referred to as extra-linguistic knowledge. In this paper, we categorize world knowledge into general knowledge and domain knowledge. The general knowledge includes commonsense knowledge \cite{davis2014representations} and prevalent knowledge. The commonsense knowledge consists of world facts, such as "Lemons are sour", or "Cows say moo", that most humans are expected to know. The prevalent knowledge exists at a particular time or place. For example, "Chinese people are used to drinking boiled water." is only known by a part of human beings; "There were eight planets in the solar system" is prevalent knowledge until it is overthrown. The domain knowledge \cite{alexander1992domain} is of a specific, specialized discipline or field, in contrast to general or domain-independent knowledge. People who have domain knowledge, are often considered specialists or experts in the field.

The bottom group of Table \ref{knowledge-datasets} shows some task examples that are used for testing world knowledge. For example, the LexGLUE \cite{DBLP:conf/acl/ChalkidisJHBAKA22} tests whether LLMs perform well in the legal domain; WikiFact \cite{DBLP:conf/acl/YasunagaLL22} is a fact completion scenario that tests language models’ factual knowledge based on Wikipedia. The input will be a partial sentence such as “The capital of France is \_”, and the output will be the continuation of the sentence such as “Paris”; TruthfulQA \cite{DBLP:conf/acl/LinHE22} comprises questions spanning numerous categories including economics, science, and law. The questions are strategically chosen so humans may also incorrectly answer them based on misconceptions and biases; language models should ideally return accurate and truthful responses; HellaSwag \cite{DBLP:conf/acl/ZellersHBFC19} tests commonsense inference and was created through adversarial filtering to synthesize wrong answers. The World knowledge competency, along with linguistic knowledge, serves as the foundation for solving different NLP tasks and is one of the core competencies of LLMs.

%???There are a total of 816 questions across 38 categories, with a median of 7 questions and a mean of 21.5 questions per category. 

%BoolQ \cite{DBLP:conf/naacl/ClarkLCK0T19} is a collection of binary yes/no questions. Each example consists of a question (Q), an excerpt from a passage (P), and an answer (A) with an explanation added for clarity.

\begin{table*}[t]
%\small
%\normalsize
%\scriptsize
\footnotesize
\begin{center} 
\begin{tabular}{l cccc }
\hline
Dataset & Knowledge Category & LLM evaluated & Task Format & Lang \\
\hline
BLiMP &  grammatical & MT-NLG;BLOOM & Classification & En \\
linguistic\_mappings & grammar/syntax & Gopher;Chinchilla;FLAN-T5;GLM;etc. & Generation & En \\
%word\_sorting & grammatical & Gopher;Chinchilla;FLAN-T5;GLM-130B;Galactica & Ranking  & En \\
minute\_mysteries\_qa & semantic & Gopher;Chinchilla;FLAN-T5;GLM;etc.  & Generation/QA  & En \\
metaphor\_boolean  & pragmatic/semantic & Gopher;Chinchilla;FLAN-T5;GLM;etc.  & Classification & En \\
\hline
%CMeEE & domain  & CPM-2 & Classification  & Ch \\
LexGLUE & domain & BLOOM &  Multiple choice & En\\
WikiFact & world & BLOOM & Generation & En\\
TruthfulQA & world & GPT-3/InstructGPT/GPT-4 & Generation & En\\
HellaSwag & commonsense & GPT-3/InstructGPT/GPT-4 & Generation & En\\
%MS MARCO& world & GPT-3/InstructGPT & ranking & En\\
\hline
\end{tabular}
\caption{Datasets that are used to evaluate the knowledge Competency of LLMs.}
\vspace{-0.5cm}
\label{knowledge-datasets}
\end{center} 
\end{table*}

\subsection{Reasoning}
%

%Reasoning, as an essential ability for complex problem-solving, can provide back-end support for various real-world applications, such as medical diagnosis, negotiation, etc. In this section, we introduce research works with comparisons and summaries and provide systematic resources to help beginners. We also discuss the potential reasons for emerging such reasoning abilities and highlight future research directions.~\cite{DBLP:journals/corr/abs-2206-04615}
% Reasoning competency is a crucial skill for solving complex problems. Moreover, from the perspective of intelligent agents, reasoning ability is also one of the core capabilities towards achieving AGI~\cite{DBLP:journals/corr/abs-2303-12712,DBLP:journals/corr/abs-2212-09597}. However, unlocking this ability in current large-scale language models often requires a substantial number of parameters, leading to an emergent state~\cite{DBLP:journals/tmlr/WeiTBRZBYBZMCHVLDF22,DBLP:journals/corr/abs-2206-04615}. Furthermore, based on the results of several common reasoning benchmarks, models still heavily rely on prompt engineering to achieve desirable outcomes in reasoning tasks~\cite{DBLP:conf/nips/Wei0SBIXCLZ22,DBLP:conf/iclr/0002WSLCNCZ23,DBLP:conf/iclr/ZhouSHWS0SCBLC23}. In this section, we decompose the reasoning competency of the model into 6 sub-capabilities, providing a comprehensive overview of existing research efforts and suggesting potential future directions.

Reasoning competency is a crucial skill for LLMs to solve complex problems. What's more, from the perspective of intelligent agents, reasoning ability is also one of the core capabilities towards achieving AGI~\cite{gpt4,DBLP:journals/corr/abs-2212-09597}. However, there remains no consensus whether LLMs can really reason, or just simply produce a larger context that increases the likelihood of correctly predicting the missing tokens~\cite{DBLP:journals/corr/abs-2302-07842}. Although "reasoning" itself may currently be an excuse of language, we can still objectively verify the reasoning performance of LLMs through various reasoning competencies. Previous methods mainly focus on the division of reasoning tasks.
\newcite{yu2023natural} divides existing evaluation tasks into three major categories, namely knowledge reasoning, symbolic reasoning, and mathematical reasoning, based on the type of logic and evidence involved in the reasoning process. \newcite{DBLP:journals/corr/abs-2303-18223} divides reasoning tasks into deductive reasoning and defeasible reasoning according to the reasoning form. In this section, we decompose the reasoning competency into 6 sub-parts from the perspective of model competency, providing a comprehensive overview of existing research efforts and suggesting potential future directions. 
And Table \ref{reasoning-datasets} presents some datasets for evaluating LLM's reasoning competency using this categorization approach.

\begin{table*}[t]
%\small
%\normalsize
%\scriptsize
\footnotesize
\begin{center} 
\begin{tabular}{l  c c c c }
\hline
Dataset & Reasoning Competency & LLM evaluated & Task Format & Lang \\
\hline
COPA &  Causal/Commonsense*  & UL2;Deberta;GLaM;GPT3;PaLM;etc. & Classification &  En \\ 
Mathematical Induction & Induction/Mathematical* &  Gopher;Chinchilla;FLAN-T5;GLM;etc. & Generation &  En \\
Synthetic Reasoning & Abduction/Deduction & HELM &  Multiple choice & En \\
SAT Analogy & Analogical & GPT-3 & Multiple choice & En \\
StrategyQA  & Multi-hop/Commonsense* & Gopher;Chinchilla;FLAN-T5;GLM;etc. & Classification & En \\
GSM8K & Mathematical* & BLOOM;LLaMA;GPT-4;MT-NLG & Generation &  En \\
ToTTo & Structured Data* & UL2 & Generation & En \\
%  & & & & En \\
% empirical\_judgments & Causal &  Gopher;Chinchilla;FLAN-T5;GLM;etc. & Multiple choice & En \\
% multistep\_arithmetic & Multi-hop/Mathematical* &  Gopher;Chinchilla;FLAN-T5;GLM;etc.& Generation &  En \\

\hline
\end{tabular}
\caption{Datasets that are used to evaluate the reasoning competency of LLMs. * represents a specific reasoning scenario.}
\vspace{-0.5cm}
\label{reasoning-datasets}
\end{center} 
\end{table*}

\subsubsection{Causal Reasoning Competency}
% 【FINISHED】
Causal reasoning competency is a highly significant cognitive ability aimed at inferring causality through the observation of cause-effect relationships~\cite{DBLP:journals/csur/VowelsCB23,DndarCoecke2022ToWE,DBLP:journals/corr/abs-2304-14827}. It enables us to comprehend and explain the relationships between events, variables, and actions, ultimately empowering us to make informed predictions and decisions~\cite{DBLP:journals/corr/abs-2305-07375}.
% Causal reasoning competency can be classified into three levels: association, intervention, and counterfactual~\cite{Chen2020CausalRA,DBLP:journals/ijautcomp/LiuWYLL22}. Firstly, the association level addresses questions like ``How does the stone will when sun shines on the stone?" by investigating the association between ``sun" and ``stone"~\cite{DBLP:journals/corr/abs-2206-04615}. The intervention level focuses on the effects of interventions. For instance, questions like ``Will I become stronger if I go to the gym every day?" require an understanding of the outcomes of specific treatments. Such questions cannot be answered solely through learning data associations. Merely observing someone who goes to the gym every day may not be sufficient evidence to conclude that going to the gym always makes one stronger, as compared to professional athletes~\cite{DBLP:conf/semeval/GordonKR12,DBLP:conf/emnlp/PontiGMLVK20}. Lastly, the counterfactual level deals with hypothetical scenarios. These questions explore the consequences when certain conditions are not fulfilled. Counterfactual reasoning aims to compare different outcomes under identical conditions, even if the antecedent of the counterfactual question is not true~\cite{pmlr-v177-mcduff22a,DBLP:journals/access/StepinACP21}.

The benchmarks Causal-TimeBank~\cite{Mirza2014AnnotatingCI}, StoryLine~\cite{DBLP:conf/acl/CaselliV17}, and MAVEN-ERE~\cite{DBLP:conf/emnlp/WangC0PWL00LLLZ22} aim to test the existence of causal relationships between two events in sentences.
COPA~\cite{DBLP:conf/SemEval/GordonKR12} and XCOPA~\cite{DBLP:conf/emnlp/PontiGMLVK20} are evaluation benchmarks for extracting causal relationships in sentences, consisting of a set of premises and possible causes or effects. Tested systems are required to apply commonsense knowledge to identify the correct answers. e-CARE~\cite{DBLP:conf/acl/DuDX0022} and CALM-Bench~\cite{DBLP:conf/eacl/DalalBA23} introduce a set of causal querying tasks to evaluate models, which include a cause and several potential effect sentences. Additionally, an annotated and interpretable causal reasoning dataset is provided for these tasks.

\subsubsection{Deduction Reasoning Competency}
% 【FINISHED】
In the era of Large Language Models (LLMs), deductive reasoning abilities serve as the foundational skills for logical reasoning~\cite{Evans2002LogicAH}. Unlike traditional rule-based deductive reasoning systems, it involves deriving specific conclusions or answers from general and universally applicable premises using given rules and logic. Specifically, it manifests as a process of Zero-Shot Chain-of-Thought utilizing given rules~\cite{DBLP:journals/corr/abs-2301-13379,DBLP:conf/nips/KojimaGRMI22}. For instance, \cite{DBLP:conf/nips/KojimaGRMI22} introduced the ``Let's think step by step" prompt technique to better evaluate the Deduction Reasoning Competency.

Current testing of this ability often intertwines with other skills and still lacks an independent evaluation on typical text~\cite{DBLP:conf/ijcai/ClarkTR20} and symbol-related~\cite{DBLP:conf/icml/WuRLBGS21} deductive datasets.
However, in general, almost all QA tasks can be explicitly evaluated for Deduction Reasoning using the Chain-of-Thought (CoT) approach. Therefore, the effectiveness of models' Deduction Reasoning Competency can be to some extent reflected by evaluating the performance of QA tasks after applying the CoT method.

\subsubsection{Induction Reasoning Competency}
% 【FINISHED】
In contrast to deductive reasoning, inductive reasoning aims to derive conclusions from specific observations to general principles~\cite{DBLP:journals/corr/abs-2212-10923,DBLP:journals/corr/abs-2209-11895}.
% Inductive reasoning involves deriving general principles or patterns from specific observations or instances. Unlike deductive reasoning, which moves from the general to the specific, it entails identifying patterns or regularities in the data and inferring general principles that can explain or predict future observations.
In recent years, a new paradigm of Induction Reasoning has been proposed by~\cite{DBLP:conf/iclr/ChengX0LNHXROZS23}, which requires models to generate general-purpose program code to solve a class of problems based on given contextual questions and a specific question. For example, \newcite{DBLP:conf/iclr/ChengX0LNHXROZS23}, \newcite{DBLP:journals/corr/abs-2305-09645} and \newcite{DBLP:journals/corr/abs-2303-08128} induced general principle-based solutions by generalizing each question into a universal executable language.

Therefore, for competency evaluation, while DEER~\cite{DBLP:journals/corr/abs-2212-10923} and Mathematical Induction (BIGBench Split~\cite{DBLP:journals/corr/abs-2206-04615}) took the first step in inductive reasoning, we still hope to establish a more systematic and comprehensive benchmark for evaluating this capability.
Recently, \newcite{bills2023language} has tested the inductive ability of GPT-4~\cite{DBLP:journals/corr/abs-2303-08774} to evaluate its effectiveness in inducing patterns that are difficult for humans to express clearly. Intriguingly, \newcite{Mankowitz2023FasterSA} used some techniques to evaluate the extent to which LLM can mine previously unknown patterns.

\subsubsection{Abduction Reasoning Competency}
% 【FINISHED】
Abduction Reasoning Competency encompasses the task of providing explanations for the output generated based on given inputs~\cite{DBLP:journals/corr/abs-2010-12896}. This form of reasoning is particularly critical in scenarios where uncertainty or incomplete information exists, enabling systems to generate hypotheses and make informed decisions based on the available evidence. Notably, the research conducted by LIREx~\cite{DBLP:conf/aaai/ZhaoV21} and STaR~\cite{Zelikman2022STaRBR} delved into the Abduction Reasoning Competency of models and demonstrated the effectiveness of rationales provided during the Abduction Reasoning process in facilitating improved learning in downstream models.

In terms of datasets within the LLM setting, 
the benchmarks HUMMINGBIRD~\cite{DBLP:conf/aaai/MathewSYBG021} and HateXplain~\cite{DBLP:conf/emnlp/HayatiKU21} require models to output word-level textual segments as explanations for sentiment classification results.
On the other hand, benchmarks such as WikiQA~\cite{DBLP:conf/emnlp/YangYM15}, HotpotQA~\cite{DBLP:conf/emnlp/Yang0ZBCSM18}, and SciFact~\cite{DBLP:conf/emnlp/WaddenLLWZCH20} provide sentence-level coarse-grained textual segments as explanations for model classification results.
ERASER~\cite{DBLP:conf/acl/DeYoungJRLXSW20} and FineIEB~\cite{DBLP:journals/corr/abs-2205-11097} provide benchmarks for evaluating Abduction Reasoning with diverse granularity explanations.  Based on previous research, Synthetic Reasoning~\cite{DBLP:journals/corr/abs-2211-09110}  provides a comprehensive evaluation of both Deduction Reasoning and Abduction Reasoning Competency.
Moreover, \newcite{DBLP:conf/eccv/HesselHPZBRSC22} introduced the first comprehensive multimodal benchmark for testing Abduction Reasoning capabilities, providing a solid foundation for future advancements in this domain.
Recently, \newcite{bills2023language} evaluate GPT-4 by observing the activation of neurons in GPT-2 and offering explanations for the GPT-2's outputs. This research avenue also presents a novel approach for exploring the future evaluation of Abduction Reasoning Competency.

\subsubsection{Analogical Reasoning Competency}
% 【FINISHED】
Analogy reasoning competency encompasses the ability of reasoning by identifying and applying similarities between diverse situations or domains. It is based on the assumption that similar cases or objects tend to exhibit common attributes or behaviors. By recognizing these similarities, analogy reasoning enables systems to transfer knowledge or experience from one context to another~\cite{DBLP:conf/emnlp/SinhaSDPH19,DBLP:conf/nips/Wei0SBIXCLZ22}. This type of reasoning plays a vital role in problem-solving, decision-making, and learning from past experiences.
A typical example is In-Context-Learning~\cite{DBLP:journals/corr/abs-2301-00234}, where the model is required to perform analogical reasoning based on given contexts, which are evaluated based on the final analogical results. 

For a better assessment and understanding of the model's analogical reasoning ability, \newcite{DBLP:conf/nips/BrownMRSKDNSSAA20} introduces SAT Analogies as a test to evaluate LLM's analogical reasoning capabilities. In recent years, Authorship Verification and ARC datasets~\cite{DBLP:journals/corr/abs-2206-04615} have also proposed evaluation benchmark that involve presenting contextual examples and requiring the model to produce induced pattern-compliant results.
However, it should be noted that In-Context Learning (ICL) can be utilized for almost all tasks, enabling the evaluation of models' Analogical Reasoning Competency to some extent through the assessment of their performance after undergoing ICL.

\subsubsection{Multi-hop Reasoning Competency}
% 【FINISHED】
Multi-hop reasoning refers to the ability to combine and integrate information from multiple sources or contexts to arrive at logical conclusions.
% It involves formulating a series of interconnected reasoning steps, where each step builds upon the information obtained from previous steps~\cite{DBLP:conf/acl/QiuXQZLZY19}. Multi-hop reasoning is particularly useful in addressing complex problems that require synthesizing information from various documents, paragraphs, or knowledge bases.
This competency of reasoning enables systems to retrieve coherent and comprehensive answers by traversing multiple pieces of information, thus performing complex tasks of information retrieval, comprehension, and reasoning~\cite{DBLP:journals/corr/abs-2212-13465,DBLP:conf/acl/QiuXQZLZY19}.

Currently, HotpotQA~\cite{DBLP:conf/emnlp/Yang0ZBCSM18} serves as a commonly used dataset for multi-hop question answering tasks. Expanding on this, \newcite{DBLP:conf/nips/YeD22} introduced a new and demanding subset that aimed to achieve a balance between accurate and inaccurate predictions using their model.
Similarly, StrategyQA~\cite{DBLP:journals/tacl/GevaKSKRB21} is another widely used benchmark for multi-hop question answering~\cite{DBLP:conf/nips/Wei0SBIXCLZ22}, where the required reasoning steps are implicit in the questions and should be inferred using strategies.

\subsubsection{Reasoning in Scenarios}
\paragraph{Commonsense Reasoning}
% 【DOING】
Commonsense reasoning is crucial for machines to achieve human-like understanding and interaction with the world in the field of machine intelligence~\cite{DBLP:journals/corr/abs-1904-01172,DBLP:conf/aaai/Bhargava022}. The ability to comprehend and apply commonsense knowledge enables machines to make accurate predictions, engage in logical reasoning, and navigate complex social situations.
% With the increase in the size of language models' parameters and training data, models have gradually deepened their understanding of the world, leading to the emergence of commonsense understanding and reasoning abilities~\cite{DBLP:conf/emnlp/WeiGLP21,DBLP:conf/acl/ZhangWLB20,DBLP:journals/tmlr/WeiTBRZBYBZMCHVLDF22}.

OpenBookQA~\cite{DBLP:conf/emnlp/MihaylovCKS18} provides a foundational test for evaluating Commonsense Reasoning abilities in the form of an open-book exam.
Building upon this, CommonsenseQA~\cite{DBLP:conf/naacl/TalmorHLB19} requires models to employ rich world knowledge for reasoning tasks.
PIQA~\cite{DBLP:conf/aaai/BiskZLGC20} introduces a dataset for testing models' understanding of physical world commonsense reasoning.
StrategyQA~\cite{DBLP:journals/tacl/GevaKSKRB21} presents a complex benchmark that requires commonsense-based multi-step/multi-hop reasoning, enabling a better exploration of the upper limits of models' Commonsense Reasoning Competency.
Currently, due to early research on LLM~\cite{DBLP:conf/nips/Wei0SBIXCLZ22}, CommonsenseQA~\cite{DBLP:conf/naacl/TalmorHLB19} remains the most widely used benchmark for commonsense reasoning.

\paragraph{Mathematical Reasoning}
% 【FINISHED】
Mathematical reasoning competency is crucial for general intelligent systems. It empowers intelligent systems with the capability of logical reasoning, problem-solving, and data manipulation and analysis, thereby facilitating the development and application of intelligent systems~\cite{DBLP:journals/corr/abs-2212-09597,DBLP:conf/acl/MishraMVSCBK22,DBLP:conf/emnlp/MishraFLTWBRTSC22}.

Early evluation studies focused on small datasets of elementary-level mathematical word problems (MWPs)~\cite{DBLP:conf/emnlp/HosseiniHEK14}, but subsequent research aimed to increase complexity and scale~\cite{DBLP:journals/corr/abs-2206-04615,DBLP:conf/nips/BrownMRSKDNSSAA20}. Furthermore, recent benchmarks~\cite{DBLP:conf/acl/MishraMVSCBK22,DBLP:conf/emnlp/MishraFLTWBRTSC22} have provided comprehensive evaluation platforms and benchmarks for mathematical reasoning abilities. GSM8K~\cite{DBLP:journals/corr/abs-2110-14168} aims to evaluate elementary school MWPs. Currently, due to early research efforts on LLMs~\cite{DBLP:conf/nips/Wei0SBIXCLZ22}, it remains the most widely used benchmark for mathematical reasoning in the LLM evaluation.
Moreover, There have been recent advancements in evaluation research that explore mathematical reasoning competency integrating external knowledge, leveraging language diversity for multilingual evaluation~\cite{DBLP:conf/iclr/ShiSF0SVCTRZ0W23}, and testing mathematical reasoning on multi-modal setting~\cite{DBLP:conf/nesy/LindstromA22}, aiming to judge the broader data reasoning capabilities of large language models (LLMs).

\paragraph{Structured Data Reasoning}
% 【FINISHED】
Structured data reasoning involves the ability to reason and derive insights and answers from structured data sources, such as structured tabular data~\cite{DBLP:journals/corr/abs-2212-09597,DBLP:journals/corr/abs-2305-13269,DBLP:conf/emnlp/XieW0ZSYWZYWZWL22}.
% It requires understanding the relationships, constraints, and even heterogeneous hierarchies within the data to perform complex queries, aggregations, and computations.
% Tasks related to structured data reasoning in the context of tabular data often involve logical operations such as filtering, sorting, joining, and grouping~\cite{DBLP:journals/corr/abs-2305-13269}, followed by the extraction of meaningful information and calculations to answer specific questions~\cite{DBLP:conf/iclr/Lu0CWZRCK23}.
% Furthermore, leveraging the strong structured data reasoning capabilities of LLMs, researchers often utilize LLMs to generate structured reasoning data to guide the training of smaller models~\cite{DBLP:conf/acl/YoranTB22}.
% Overall, the optimization of structured data reasoning capabilities in LLMs is still in its early stages.
% Recent evaluation research efforts have primarily focused on developing benchmark datasets~\cite{DBLP:journals/corr/abs-2212-09597}, utilizing prompts and tools to enhance structured data reasoning in machine intelligence~\cite{DBLP:conf/iclr/Lu0CWZRCK23,DBLP:journals/corr/abs-2212-08607}.

WikiSQL~\cite{DBLP:journals/corr/abs-1709-00103} and WikiTQ~\cite{DBLP:conf/acl/PasupatL15} provide tables as input and answer questions based on the additional input of questions. HybridQA~\cite{DBLP:conf/emnlp/ChenZCXWW20} and MultiModalQA~\cite{DBLP:conf/iclr/TalmorYCLWAIHB21} propose benchmarks for hybrid Structure Reasoning by combining structured table inputs with text (and even other modalities). Similarly, MultiWoZ~\cite{DBLP:conf/emnlp/BudzianowskiWTC18}, KVRET~\cite{DBLP:conf/sigdial/EricKCM17} and SQA~\cite{DBLP:conf/acl/IyyerYC17} integrate table data into task-oriented dialogue systems to generate more complex structures and output dialog-related classifications. Unlike traditional QA, FeTaQA~\cite{Nan2021FeTaQAFT} requires free-form answers instead of extracting answer spans from passages. 
ToTTo~\cite{DBLP:conf/emnlp/ParikhWGFDYD20} introduces an open-domain English table-to-text dataset for Structured Data Reasoning.
Additionally, benchmarks such as TabFact~\cite{DBLP:conf/iclr/ChenWCZWLZW20} and FEVEROUS~\cite{Aly2021TheFE} evaluate whether model statements are consistent with facts mentioned in structured data. In recent years, with a deeper focus on testing models' mathematical abilities, TabMWP~\cite{DBLP:conf/iclr/Lu0CWZRCK23} introduces a grade-level dataset of table-based mathematical word problems that require mathematical reasoning using both text and table data.
\subsection{Reliability}
Reliability measures to what extent a human can trust the contents generated by a LLM. It is of vital importance for the deployment and usability of the LLM, and attracts tons of concerns along with the rapid
 and astonishing development of recent LLMs \cite{DBLP:journals/corr/abs-2112-04359,DBLP:conf/naacl/0002WY22,DBLP:journals/csur/JiLFYSXIBMF23,zhuo2023red}. Lots of concepts are closely related to reliability under the context of LLM, including but not limited to hallucination, truthfulness, factuality, honesty, calibration, robustness, interpretability \cite{Lee2018HallucinationsIN,belinkov-etal-2020-interpretability,DBLP:journals/corr/abs-2110-06674,DBLP:journals/tacl/MielkeSDB22,DBLP:conf/acl/LinHE22}. Reliability also overlaps with the safety and generalization of a LLM \cite{DBLP:journals/corr/abs-2112-04359}. In this section, we will give an overview of two most concerned directions: Hallucination, Uncertainty and Calibration.
 \subsubsection{Hallucination}
 
 Hallucination is a term often used to describe LLM's falsehoods, which is the opposite side of truthfulness or factuality \cite{DBLP:journals/csur/JiLFYSXIBMF23,DBLP:journals/corr/abs-2303-08774,gpt4}. Hallucination is always categorized into intrinsic (close domain) hallucination and extrinsic (open domain) hallucination \cite{DBLP:journals/csur/JiLFYSXIBMF23,DBLP:journals/corr/abs-2303-08774}. Intrinsic hallucination refers to the unfaithfulness of the model output to a given context, while extrinsic hallucination refers to the untruthful contents about the world generated by the model without reference to a given source.
 
 Early research on hallucination mainly focused on the intrinsic hallucination and lots of interesting metrics were proposed to evaluate the intrinsic hallucination level of a PTM \cite{DBLP:journals/csur/JiLFYSXIBMF23}. However, \newcite{DBLP:journals/corr/abs-2302-04023} claimed that intrinsic hallucination was barely found after conducting a comprehensive analysis of ChatGPT's responses. Hence for LLM, the extrinsic hallucination is of the greatest concern. To evaluate the extrinsic hallucination potential of a LLM, a common practice is to leverage knowledge-intensive tasks such as Factual Question Answering \cite{DBLP:conf/acl/JoshiCWZ17,zheng2023does} or Knowledge-grounded Dialogue \cite{DBLP:conf/iclr/DinanRSFAW19,DBLP:conf/emnlp/DasSS22}. TruthfulQA \cite{DBLP:conf/acl/LinHE22} is the most popular dataset used to quantify hallucination level of a LLM. This dataset is adversarially constructed to exploit the weakness of LLM, which contained 817 questions that span 38 categories. \newcite{DBLP:journals/corr/abs-2303-08774} leveraged real-world data flagged as non-factual to construct an adversarial dataset to test GPT-4's hallucination potential. BIG-bench \cite{DBLP:journals/corr/abs-2206-04615}, a famous benchmark to evaluate LLM's capabilities, also contains many sub-tasks on factual correctness including TruthfulQA. Although most of these tasks are multiple choices or classification in a fact verification\cite{DBLP:conf/naacl/ThorneVCM18} manner, they are closely associated with truthfulness and can be regarded as a generalized hallucination evaluation.
 
 \subsubsection{Uncertainty and Calibration}
 A reliable and trustworthy Language model must have the capability to accurately articulate its level of confidence over its response, which requires the model to be aware of its uncertainty. A model that can precisely measure its own uncertainty is sometimes called self-aware, honesty or known-unknown \cite{DBLP:journals/corr/abs-2207-05221,DBLP:journals/corr/abs-2305-18153}. In general deep learning applications, calibration concerns about the uncertainty estimation of a classifier. Output probability from a well-calibrated classifier are supposed to be consistent with the empirical accuracy in real world \cite{DBLP:conf/aistats/VaicenaviciusWA19}. HELM \cite{DBLP:journals/corr/abs-2211-09110} treated calibration as one of general metrics and comprehensively evaluated the calibration degree of many prevailing models on multiple choice and classification tasks. \cite{DBLP:journals/corr/abs-2303-08774} also showed that GPT-4 before RLHF was well-calibrated on multiple choice tasks, although the decent calibration degree was compromised significantly by post-training.

 when it comes to free-form generation, it's a different story. \newcite{DBLP:conf/iclr/KuhnGF23} pointed out that semantic nature of language and intractable output space guaranteed the uniqueness of free-form generation. They proposed an algorithm to cluster model outputs and then estimate the model uncertainty. \newcite{DBLP:journals/tacl/MielkeSDB22} claimed that models always express confidence over incorrect answers and proposed the notion of linguistic calibration, which teached models to verbally express uncertainty rather than estimating a probability. \newcite{DBLP:journals/tmlr/LinHE22} trained models to directly generate predicted uncertainty probability in natural language. \newcite{DBLP:journals/corr/abs-2305-18153} proposed the SelfAware dataset which contains unanswerable questions and used the accuracy of model rejection as a measure of uncertainty.
\subsection{Safety}
As the LLMs rapidly penetrate into  the manufactural and interactive activities of human society, such as LLM-based poem-template generators and chatting robots, the safety concerns for LLMs gain much attention nowadays. The rationales of LLMs are statistics-based, and this inherent stochasticity brings limitations and underlying risks, which deeply affect the real-world deployment of LLMs. Some datasets are proposed to evaluated the safety of LLMs (Table \ref{safety-datasets}), however, the corresponding validity and authority of the safety judgement are inadequate as the current evaluative dimensions are not sufficient \cite{UnderstandingAbuse,DBLP:journals/corr/abs-2112-04359} and the perception of safety is highly subjective \cite{Offensiveaggressive,DBLP:journals/corr/abs-2112-04359}. To this end, based on our survey on relevant papers, we propose a comprehensive perspective on the safety competency of LLMs, ranging from harmful contents to the ethical consideration, to inspire the further developments towards the techniques and evaluations of LLMs safety.

\subsubsection{Harmfulness}
The harmful contents include the offensive language or others that have the explicit harm towards the specific object, such content that has been widely discussed. However, there is not a unified definition of the constitution of harmful contents, based on our surveys, we conclude the relevant themes into five aspects, including offensiveness, violence, crime, sexual-explicit, and unauthorized expertise. Many researches focus on the language detection for the outputs of LLMs to ensure the harmlessness \cite{ExMachina:,AutomatedHS,Predictingthetypeand,BuilditBreakitFixit}, while other techniques are proposed to stimulate LLMs to generate safe outputs directly \cite{gedi,appdia}. For the unauthorized expertise, a general LLM should avoid any unauthorized expertise before the establishment of accountability system \cite{onthesafetyof}, which involves the psychological orientation and any medical advice. Besides, the impact of conversation context on safety gains more attention recently, as a results, detective and generative algorithms base on the context are proposed successively \cite{BuilditBreakitFixit,justsayno,safetykit}. RealToxicityPrompts \cite{realtoxicityprompts} is a dataset derived from English web texts, where prompts are automatically truncated from sentences classified as toxicity from a widely-used toxicity classifier. RealToxicityPrompts consists of 100K natural prompts, with average 11.7 tokens in length. BAD \cite{BotAdversarial} is a dataset collected by the human-in-the-loop strategy, where crowdworkers are ask to prob harmful model outputs. BAD consist of 5k conversations with around 70k utterances in total, which could be used in both non-adversarially and adversarially testing the model weakness.

\begin{table*}[t]
\footnotesize
\begin{center} 
\begin{tabular}{l  c c c c }
\hline
Dataset & Safety Category & LLM evaluated & Task Format & Lang \\
\hline
RealToxicityPrompts & Harmful Contents & InstructGPT;LLaMA;Flan-PaLM;GPT-4;BLOOM & Generation &  En \\ 
BAD & Harmful Contents & - & Generation &  En \\ 
CrowS-Pairs & Social Bias & LLaMA;MT-NLG;InstructGPT;Pythia & Generatio &  En \\ 
French CrowS-Pairs & Social Bias & MT-NLG & Generation &  Fr \\ 
StereoSet  & Social Bias & - & Multiple choice &  En \\ 

\hline
\end{tabular}
\caption{Datasets used to evaluate the safety competency of LLMs.}
\vspace{-0.5cm}
\label{safety-datasets}
\end{center} 
\end{table*}

\subsubsection{Unfairness and Social Bias}
Unfairness and social bias present more covertly and widely for LLMs. Following the previous studies, we conclude that social bias is an inherent characteristic of a LLM, which mainly embody in the distribution difference of a LLM in language selection based on different demographic groups. Compared to the social bias, unfairness is the external form, which reflected in the output performance of specific tasks, for example, the African American English (AAE) is frequently mis-classified as the offensive language by some language detector \cite{measuringgeopraphic}. However, issues of unfairness and social bias are inevitable as they are widely distributed in human languages, and LLMs are required to memorize language as accurately as possible in the training stage \cite{DBLP:journals/corr/abs-2112-04359}. With respect to evaluate this important aspect, CrowS-Pairs \cite{crowspairs} is benchmark proposed to evaluating social bias. There are 1508 examples in CrowS-Pairs that involves nine types of social bias, like gender, race, and Nationality. StereoSet \cite{stereoset} is a dataset that could be used to evaluate social bias level in both word-level and sentence level, which examples are in four domains: race, gender,religion, and profession. For the StereoSet, the bias level is computed by the difference between model generation probabilities of biased and anti-biased sentence.

\subsubsection{Others}
As current algorithms for model safety based on the human perception, there is still no golden standardized judgement for LLMs to refer to, especially when a judgement is highly various across societies. It is necessary to align LLMs with the morality, ethics, and values of human society. More and more works focus on reifying this abstract concept into textual data recently, for example, \newcite{SocialBiasFrames} proposal an implicit reasoning frame to explain the underlying harm of the target language. Besides, other works leverage rule-of-thumb (RoT) annotations of texts to support the judgement \cite{SocialChemistry101,themoralintegrity}. However, current works in this area are neonatal, and we could expect more related works in the future.

Besides, we are also concerned about the privacy and political risks of LLMs. Since the LLMs are trained on vast corpus collected from books, conversations, web texts and so on, the privacy safety of LLMs arouses people's concern. These training texts might contain the private or sensitive information such as personal physical information, home address, etc. Many studies indicate LLMs are brittle under attacks, leaking the sensitive information unintentionally \cite{ExtractingTrainingData,YouDontKnowMy}. Therefore, it is essential to test the privacy protection ability of a LLM. Moreover, the politics ignorance is also intractable for a LLM. The politics-related risk mainly stems from the composition of the training corpus. Texts in the corpus are derived from different language and social environments (usually the larger the more diversified), and different countries have different political prudence and stance, which brings additional risks to the wide deployment of a LM.

\section{Future Directions}
In this section, we outline some other competencies that are important for evaluating LLMs.

\subsection{Sentiment}
It is crucial to equip LLMs with the ability to understand and generate sentiments. As an indispensable factor in human life, sentiments are widely present in daily chats, social media posts, customer reviews, and news articles \cite{SC_2015}. Through the comprehensive research and high-level summary of the literature related to sentiments, we introduce the sentiment competency of LLMs in two aspects: sentiment understand and sentiment generation.

% In this section, we introduce the sentiment competency of LLMs in two aspects: sentiment understand and sentiment generation.

\subsubsection{Sentiment Understanding}
Sentiment understand mainly involves the understanding of opinions, sentiments and emotions in the text \cite{SC_2015}. Representative tasks that reflect this competency include sentiment classification (SC), aspect-based sentiment analysis (ABSA), and multifaceted analysis of subjective texts (MAST). SC aims at assigning pre-defined sentiment classes to given texts. The typical datasets include IMDB \cite{dataset_imdb}, SST \cite{dataset_SST}, Twitter \cite{dataset_twitter_SSC}, Yelp \cite{dataset_yelp5}. ABSA focuses on identifying the sentiments of specific aspects in a sentence \cite{absa_survey}, and the most widely used datasets are the SemEval series \cite{SemEval-2014_Task4,SemEval-2015_Task12,semeval2016_Task5}.
MAST are tasks that involve the finer-grained and broader range of human subjective feelings (emotions \cite{emotion}, stance \cite{stance}, hate \cite{hate}, irony \cite{iron}, offensive \cite{offensive}, etc.) \cite{MAST-ref}. Given that MAST includes a wide range of tasks, the datasets are not listed here in detail. Among them, the commonly used evaluation metrics for the above tasks are accuracy and F1 score (micro or macro). Some preliminary empirical studies \cite{sc_llm_1,sc_chatgpt_1} indicate that LLMs can significantly improve performance on these tasks in few-shot learning settings. LLMs have the potential to be a general solution without designing different models for various tasks. Therefore, the sentiment understand competency of different LLMs deserves comprehensive exploration and empirical evaluation. To evaluate the performance of this competency, we can utilize multiple domain-specific datasets or choose the comprehensive benchmark \cite{DBLP:journals/corr/abs-2206-04615,DBLP:journals/corr/abs-2211-09110}.

\subsubsection{Sentiment Generation}
We categorize sentiment generation into two manifestations. One is to generate text that contains sentiments, and the other is to generate text that elicits sentiments. The former requires specifying the desired sentiment, and the latter requires a combination of commonsense knowledge \cite{conceptnet,Atomic} or theory of mind \cite{tom}. A classic application scenario is in open-domain dialogue, specifically, emotional dialogue \cite{ecm}, empathetic dialogue \cite{emp}, and emotional support conversation \cite{ESC}. To measure the quality of the generated text, it is necessary to employ both automatic metrics (such as sentiment accuracy, BLEU \cite{bleu}, perplexity) and human evaluations (human ratings or preference tests). Currently, no work has comprehensively explored this aspect, but it is an essential path towards artificial general intelligence (AGI) \cite{gpt4}.
\subsection{Planning}
Planning is the thinking before the actions take place. Given a specific goal, planning is the process to decide the means to achieve the goal. There're few works \cite{DBLP:journals/corr/abs-2305-15771,DBLP:journals/corr/abs-2206-10498,DBLP:journals/corr/abs-2305-16151,DBLP:conf/icml/HuangAPM22} that look at the planning ability of LLMs. Some of them focus on commonsense areas \cite{DBLP:conf/icml/HuangAPM22} like wedding or menu planning. Others adopted automated planning problems, formal language translators, and verifiers to automatically evaluate LLMs' competency\cite{DBLP:journals/corr/abs-2305-15771}. With PDDL \footnote{Planning Domain Definition Language, a formal language used to describe classical planning problems.} represented problem descriptions and the translation of such problems into text and back, LLMs can thus sequence a series of actions to reach the planning goal. Whether the planning purpose is achieved can be easily verified via automatic verifiers. Possessing web-scale knowledge, LLMs have great potential for executing planning tasks or assisting planners. 

\subsection{Code}
Coding competency is one of the advanced abilities of LLMs. LLMs with this competency can not only perform program synthesis but also possess the potential of self-evolving. Technically, all of the tasks involved with code like code generation and code understanding need this competency. In oracle manual evaluation, prominent LLMs like ChatGPT are capable of up to 15 ubiquitous software engineering tasks and perform well in most of them \cite{DBLP:journals/corr/abs-2305-16837}. The most explored evaluation task in coding competency would be program synthesis, where program description and function signature are given for its code implementation. One of the most pioneering benchmarks in program synthesis, HUMANEVAL \cite{DBLP:journals/corr/abs-2107-03374}, consists of 164 pairs of human-generated docstrings and the associated unit tests to test the functional correctness of model generation. However, with the worry of insufficient testing and the imprecise problem description \cite{DBLP:journals/corr/abs-2305-01210}, existing LLM-for-code benchmarks still have lots of room for improvement.
%\subsection{multi-modal}
%\subsection{multi-lingual}

\section{Conclusion}
This survey provides a comprehensive review of various literature for the evaluation of LLMs. We aggregate different works with their intended competencies. Some of the competencies(reasoning, knowledge) already have holistic evaluation benchmarks, while others(planning, coding) still face disparate challenges. The goal of this paper is to comb the numerous work concerning LLMs' evaluation through the lens of the core competencies test. Lighten the cognitive load for assimilating numerous evaluation works due to the various functions of LLMs. In doing so, we have also identified the challenge faced by each competency, looking forward to alleviating it in the future.

\section*{Acknowledgements}
We want to thank Yuanxing Liu, Xuesong Wang, Mengzhou Sun, Runze Liu, Yuhang Gou, Shuhan Zhou, Yifan Chen, Ruiyu Xiao, Xinyu Li, Yuchi Zhang, Yang Wang, Jiahang Han, Wenqi Ding, and Xinpeng Liu for their priceless help with the initial dataset investigation process. 

\bibliographystyle{ccl}
\bibliography{ccl}

\begin{thebibliography}{}

\bibitem[\protect\citename{Alexander}1992]{alexander1992domain}
Patricia~A Alexander.
\newblock 1992.
\newblock Domain knowledge: Evolving themes and emerging concerns.
\newblock {\em Educational psychologist}, 27(1):33--51.

\bibitem[\protect\citename{Aly \bgroup et al.\egroup }2021]{Aly2021TheFE}
Rami Aly, Zhijiang Guo, M.~Schlichtkrull, James Thorne, Andreas Vlachos,
  Christos Christodoulopoulos, Oana Cocarascu, and Arpit Mittal.
\newblock 2021.
\newblock The fact extraction and verification over unstructured and structured
  information (feverous) shared task.
\newblock {\em Proceedings of the Fourth Workshop on Fact Extraction and
  VERification (FEVER)}.

\bibitem[\protect\citename{Atwell \bgroup et al.\egroup }2022]{appdia}
Katherine Atwell, Sabit Hassan, and Malihe Alikhani.
\newblock 2022.
\newblock {APPDIA}: A discourse-aware transformer-based style transfer model
  for offensive social media conversations.
\newblock In {\em Proceedings of the 29th International Conference on
  Computational Linguistics}, pages 6063--6074, Gyeongju, Republic of Korea,
  October. International Committee on Computational Linguistics.

\bibitem[\protect\citename{Austin}1975]{austin1975things}
John~Langshaw Austin.
\newblock 1975.
\newblock {\em How to do things with words}, volume~88.
\newblock Oxford university press.

\bibitem[\protect\citename{Baheti \bgroup et al.\egroup }2021]{justsayno}
Ashutosh Baheti, Maarten Sap, Alan Ritter, and Mark Riedl.
\newblock 2021.
\newblock Just say no: Analyzing the stance of neural dialogue generation in
  offensive contexts.
\newblock In {\em Proceedings of the 2021 Conference on Empirical Methods in
  Natural Language Processing}, pages 4846--4862, Online and Punta Cana,
  Dominican Republic, November. Association for Computational Linguistics.

\bibitem[\protect\citename{Bang \bgroup et al.\egroup
  }2023]{DBLP:journals/corr/abs-2302-04023}
Yejin Bang, Samuel Cahyawijaya, Nayeon Lee, Wenliang Dai, Dan Su, Bryan Wilie,
  Holy Lovenia, Ziwei Ji, Tiezheng Yu, Willy Chung, Quyet~V. Do, Yan Xu, and
  Pascale Fung.
\newblock 2023.
\newblock A multitask, multilingual, multimodal evaluation of chatgpt on
  reasoning, hallucination, and interactivity.
\newblock {\em CoRR}, abs/2302.04023.

\bibitem[\protect\citename{Belinkov \bgroup et al.\egroup
  }2020]{belinkov-etal-2020-interpretability}
Yonatan Belinkov, Sebastian Gehrmann, and Ellie Pavlick.
\newblock 2020.
\newblock Interpretability and analysis in neural {NLP}.
\newblock In {\em Proceedings of the 58th Annual Meeting of the Association for
  Computational Linguistics: Tutorial Abstracts}, pages 1--5, Online, July.
  Association for Computational Linguistics.

\bibitem[\protect\citename{Bhargava and Ng}2022]{DBLP:conf/aaai/Bhargava022}
Prajjwal Bhargava and Vincent Ng.
\newblock 2022.
\newblock Commonsense knowledge reasoning and generation with pre-trained
  language models: {A} survey.
\newblock In {\em Thirty-Sixth {AAAI} Conference on Artificial Intelligence,
  {AAAI} 2022, Thirty-Fourth Conference on Innovative Applications of
  Artificial Intelligence, {IAAI} 2022, The Twelveth Symposium on Educational
  Advances in Artificial Intelligence, {EAAI} 2022 Virtual Event, February 22 -
  March 1, 2022}, pages 12317--12325. {AAAI} Press.

\bibitem[\protect\citename{Bills \bgroup et al.\egroup
  }2023]{bills2023language}
Steven Bills, Nick Cammarata, Dan Mossing, Henk Tillman, Leo Gao, Gabriel Goh,
  Ilya Sutskever, Jan Leike, Jeff Wu, and William Saunders.
\newblock 2023.
\newblock Language models can explain neurons in language models.
\newblock
  \url{https://openaipublic.blob.core.windows.net/neuron-explainer/paper/index.html}.

\bibitem[\protect\citename{Bisk \bgroup et al.\egroup
  }2020]{DBLP:conf/aaai/BiskZLGC20}
Yonatan Bisk, Rowan Zellers, Ronan~Le Bras, Jianfeng Gao, and Yejin Choi.
\newblock 2020.
\newblock {PIQA:} reasoning about physical commonsense in natural language.
\newblock In {\em The Thirty-Fourth {AAAI} Conference on Artificial
  Intelligence, {AAAI} 2020, The Thirty-Second Innovative Applications of
  Artificial Intelligence Conference, {IAAI} 2020, The Tenth {AAAI} Symposium
  on Educational Advances in Artificial Intelligence, {EAAI} 2020, New York,
  NY, USA, February 7-12, 2020}, pages 7432--7439. {AAAI} Press.

\bibitem[\protect\citename{Brown \bgroup et al.\egroup
  }2020]{DBLP:conf/nips/BrownMRSKDNSSAA20}
Tom~B. Brown, Benjamin Mann, Nick Ryder, Melanie Subbiah, Jared Kaplan,
  Prafulla Dhariwal, Arvind Neelakantan, Pranav Shyam, Girish Sastry, Amanda
  Askell, Sandhini Agarwal, Ariel Herbert{-}Voss, Gretchen Krueger, Tom
  Henighan, Rewon Child, Aditya Ramesh, Daniel~M. Ziegler, Jeffrey Wu, Clemens
  Winter, Christopher Hesse, Mark Chen, Eric Sigler, Mateusz Litwin, Scott
  Gray, Benjamin Chess, Jack Clark, Christopher Berner, Sam McCandlish, Alec
  Radford, Ilya Sutskever, and Dario Amodei.
\newblock 2020.
\newblock Language models are few-shot learners.
\newblock In Hugo Larochelle, Marc'Aurelio Ranzato, Raia Hadsell,
  Maria{-}Florina Balcan, and Hsuan{-}Tien Lin, editors, {\em Advances in
  Neural Information Processing Systems 33: Annual Conference on Neural
  Information Processing Systems 2020, NeurIPS 2020, December 6-12, 2020,
  virtual}.

\bibitem[\protect\citename{Bubeck \bgroup et al.\egroup }2023]{gpt4}
S{\'{e}}bastien Bubeck, Varun Chandrasekaran, Ronen Eldan, Johannes Gehrke,
  Eric Horvitz, Ece Kamar, Peter Lee, Yin~Tat Lee, Yuanzhi Li, Scott~M.
  Lundberg, Harsha Nori, Hamid Palangi, Marco~T{\'{u}}lio Ribeiro, and
  Yi~Zhang.
\newblock 2023.
\newblock Sparks of artificial general intelligence: Early experiments with
  {GPT-4}.
\newblock {\em CoRR}, abs/2303.12712.

\bibitem[\protect\citename{Budzianowski \bgroup et al.\egroup
  }2018]{DBLP:conf/emnlp/BudzianowskiWTC18}
Pawel Budzianowski, Tsung{-}Hsien Wen, Bo{-}Hsiang Tseng, I{\~{n}}igo
  Casanueva, Stefan Ultes, Osman Ramadan, and Milica Gasic.
\newblock 2018.
\newblock Multiwoz - {A} large-scale multi-domain wizard-of-oz dataset for
  task-oriented dialogue modelling.
\newblock In Ellen Riloff, David Chiang, Julia Hockenmaier, and Jun'ichi
  Tsujii, editors, {\em Proceedings of the 2018 Conference on Empirical Methods
  in Natural Language Processing, Brussels, Belgium, October 31 - November 4,
  2018}, pages 5016--5026. Association for Computational Linguistics.

\bibitem[\protect\citename{Carlini \bgroup et al.\egroup
  }2020]{ExtractingTrainingData}
Nicholas Carlini, Florian Tram{\`{e}}r, Eric Wallace, Matthew Jagielski, Ariel
  Herbert{-}Voss, Katherine Lee, Adam Roberts, Tom~B. Brown, Dawn Song,
  {\'{U}}lfar Erlingsson, Alina Oprea, and Colin Raffel.
\newblock 2020.
\newblock Extracting training data from large language models.
\newblock {\em CoRR}, abs/2012.07805.

\bibitem[\protect\citename{Caselli and Vossen}2017]{DBLP:conf/acl/CaselliV17}
Tommaso Caselli and Piek Vossen.
\newblock 2017.
\newblock The event storyline corpus: {A} new benchmark for causal and temporal
  relation extraction.
\newblock In Tommaso Caselli, Ben Miller, Marieke van Erp, Piek Vossen, Martha
  Palmer, Eduard~H. Hovy, Teruko Mitamura, and David Caswell, editors, {\em
  Proceedings of the Events and Stories in the News Workshop@ACL 2017,
  Vancouver, Canada, August 4, 2017}, pages 77--86. Association for
  Computational Linguistics.

\bibitem[\protect\citename{Chalkidis \bgroup et al.\egroup
  }2022]{DBLP:conf/acl/ChalkidisJHBAKA22}
Ilias Chalkidis, Abhik Jana, Dirk Hartung, Michael J.~Bommarito II, Ion
  Androutsopoulos, Daniel~Martin Katz, and Nikolaos Aletras.
\newblock 2022.
\newblock Lexglue: {A} benchmark dataset for legal language understanding in
  english.
\newblock In Smaranda Muresan, Preslav Nakov, and Aline Villavicencio, editors,
  {\em Proceedings of the 60th Annual Meeting of the Association for
  Computational Linguistics (Volume 1: Long Papers), {ACL} 2022, Dublin,
  Ireland, May 22-27, 2022}, pages 4310--4330. Association for Computational
  Linguistics.

\bibitem[\protect\citename{Chan \bgroup et al.\egroup
  }2023]{DBLP:journals/corr/abs-2304-14827}
Chunkit Chan, Jiayang Cheng, Weiqi Wang, Yuxin Jiang, Tianqing Fang, Xin Liu,
  and Yangqiu Song.
\newblock 2023.
\newblock Chatgpt evaluation on sentence level relations: {A} focus on
  temporal, causal, and discourse relations.
\newblock {\em CoRR}, abs/2304.14827.

\bibitem[\protect\citename{Chen \bgroup et al.\egroup
  }2020a]{DBLP:conf/iclr/ChenWCZWLZW20}
Wenhu Chen, Hongmin Wang, Jianshu Chen, Yunkai Zhang, Hong Wang, Shiyang Li,
  Xiyou Zhou, and William~Yang Wang.
\newblock 2020a.
\newblock Tabfact: {A} large-scale dataset for table-based fact verification.
\newblock In {\em 8th International Conference on Learning Representations,
  {ICLR} 2020, Addis Ababa, Ethiopia, April 26-30, 2020}. OpenReview.net.

\bibitem[\protect\citename{Chen \bgroup et al.\egroup
  }2020b]{DBLP:conf/emnlp/ChenZCXWW20}
Wenhu Chen, Hanwen Zha, Zhiyu Chen, Wenhan Xiong, Hong Wang, and William~Yang
  Wang.
\newblock 2020b.
\newblock Hybridqa: {A} dataset of multi-hop question answering over tabular
  and textual data.
\newblock In Trevor Cohn, Yulan He, and Yang Liu, editors, {\em Findings of the
  Association for Computational Linguistics: {EMNLP} 2020, Online Event, 16-20
  November 2020}, volume {EMNLP} 2020 of {\em Findings of {ACL}}, pages
  1026--1036. Association for Computational Linguistics.

\bibitem[\protect\citename{Chen \bgroup et al.\egroup
  }2021]{DBLP:journals/corr/abs-2107-03374}
Mark Chen, Jerry Tworek, Heewoo Jun, Qiming Yuan, Henrique~Pond{\'{e}}
  de~Oliveira~Pinto, Jared Kaplan, Harrison Edwards, Yuri Burda, Nicholas
  Joseph, Greg Brockman, Alex Ray, Raul Puri, Gretchen Krueger, Michael Petrov,
  Heidy Khlaaf, Girish Sastry, Pamela Mishkin, Brooke Chan, Scott Gray, Nick
  Ryder, Mikhail Pavlov, Alethea Power, Lukasz Kaiser, Mohammad Bavarian,
  Clemens Winter, Philippe Tillet, Felipe~Petroski Such, Dave Cummings,
  Matthias Plappert, Fotios Chantzis, Elizabeth Barnes, Ariel Herbert{-}Voss,
  William~Hebgen Guss, Alex Nichol, Alex Paino, Nikolas Tezak, Jie Tang, Igor
  Babuschkin, Suchir Balaji, Shantanu Jain, William Saunders, Christopher
  Hesse, Andrew~N. Carr, Jan Leike, Joshua Achiam, Vedant Misra, Evan Morikawa,
  Alec Radford, Matthew Knight, Miles Brundage, Mira Murati, Katie Mayer, Peter
  Welinder, Bob McGrew, Dario Amodei, Sam McCandlish, Ilya Sutskever, and
  Wojciech Zaremba.
\newblock 2021.
\newblock Evaluating large language models trained on code.
\newblock {\em CoRR}, abs/2107.03374.

\bibitem[\protect\citename{Cheng \bgroup et al.\egroup
  }2023]{DBLP:conf/iclr/ChengX0LNHXROZS23}
Zhoujun Cheng, Tianbao Xie, Peng Shi, Chengzu Li, Rahul Nadkarni, Yushi Hu,
  Caiming Xiong, Dragomir Radev, Mari Ostendorf, Luke Zettlemoyer, Noah~A.
  Smith, and Tao Yu.
\newblock 2023.
\newblock Binding language models in symbolic languages.
\newblock In {\em The Eleventh International Conference on Learning
  Representations, {ICLR} 2023, Kigali, Rwanda, May 1-5, 2023}. OpenReview.net.

\bibitem[\protect\citename{Clark \bgroup et al.\egroup
  }2020]{DBLP:conf/ijcai/ClarkTR20}
Peter Clark, Oyvind Tafjord, and Kyle Richardson.
\newblock 2020.
\newblock Transformers as soft reasoners over language.
\newblock In Christian Bessiere, editor, {\em Proceedings of the Twenty-Ninth
  International Joint Conference on Artificial Intelligence, {IJCAI} 2020},
  pages 3882--3890. ijcai.org.

\bibitem[\protect\citename{Cobbe \bgroup et al.\egroup
  }2021]{DBLP:journals/corr/abs-2110-14168}
Karl Cobbe, Vineet Kosaraju, Mohammad Bavarian, Mark Chen, Heewoo Jun, Lukasz
  Kaiser, Matthias Plappert, Jerry Tworek, Jacob Hilton, Reiichiro Nakano,
  Christopher Hesse, and John Schulman.
\newblock 2021.
\newblock Training verifiers to solve math word problems.
\newblock {\em CoRR}, abs/2110.14168.

\bibitem[\protect\citename{Dalal \bgroup et al.\egroup
  }2023]{DBLP:conf/eacl/DalalBA23}
Dhairya Dalal, Paul Buitelaar, and Mihael Arcan.
\newblock 2023.
\newblock Calm-bench: {A} multi-task benchmark for evaluating causality-aware
  language models.
\newblock In Andreas Vlachos and Isabelle Augenstein, editors, {\em Findings of
  the Association for Computational Linguistics: {EACL} 2023, Dubrovnik,
  Croatia, May 2-6, 2023}, pages 296--311. Association for Computational
  Linguistics.

\bibitem[\protect\citename{Das \bgroup et al.\egroup
  }2022]{DBLP:conf/emnlp/DasSS22}
Souvik Das, Sougata Saha, and Rohini~K. Srihari.
\newblock 2022.
\newblock Diving deep into modes of fact hallucinations in dialogue systems.
\newblock In Yoav Goldberg, Zornitsa Kozareva, and Yue Zhang, editors, {\em
  Findings of the Association for Computational Linguistics: {EMNLP} 2022, Abu
  Dhabi, United Arab Emirates, December 7-11, 2022}, pages 684--699.
  Association for Computational Linguistics.

\bibitem[\protect\citename{Davidson \bgroup et al.\egroup }2017]{AutomatedHS}
Thomas Davidson, Dana Warmsley, Michael~W. Macy, and Ingmar Weber.
\newblock 2017.
\newblock Automated hate speech detection and the problem of offensive
  language.
\newblock In {\em International Conference on Web and Social Media}.

\bibitem[\protect\citename{Davis}2014]{davis2014representations}
Ernest Davis.
\newblock 2014.
\newblock {\em Representations of commonsense knowledge}.
\newblock Morgan Kaufmann.

\bibitem[\protect\citename{Day \bgroup et al.\egroup }1998]{day1998extensive}
Richard~R Day, Julian Bamford, Willy~A Renandya, George~M Jacobs, and Vivienne
  Wai-Sze Yu.
\newblock 1998.
\newblock Extensive reading in the second language classroom.
\newblock {\em RELC Journal}, 29(2):187--191.

\bibitem[\protect\citename{DeYoung \bgroup et al.\egroup
  }2020]{DBLP:conf/acl/DeYoungJRLXSW20}
Jay DeYoung, Sarthak Jain, Nazneen~Fatema Rajani, Eric Lehman, Caiming Xiong,
  Richard Socher, and Byron~C. Wallace.
\newblock 2020.
\newblock {ERASER:} {A} benchmark to evaluate rationalized {NLP} models.
\newblock In Dan Jurafsky, Joyce Chai, Natalie Schluter, and Joel~R. Tetreault,
  editors, {\em Proceedings of the 58th Annual Meeting of the Association for
  Computational Linguistics, {ACL} 2020, Online, July 5-10, 2020}, pages
  4443--4458. Association for Computational Linguistics.

\bibitem[\protect\citename{Dinan \bgroup et al.\egroup
  }2019a]{BuilditBreakitFixit}
Emily Dinan, Samuel Humeau, Bharath Chintagunta, and Jason Weston.
\newblock 2019a.
\newblock Build it break it fix it for dialogue safety: Robustness from
  adversarial human attack.
\newblock In {\em Proceedings of the 2019 Conference on Empirical Methods in
  Natural Language Processing and the 9th International Joint Conference on
  Natural Language Processing (EMNLP-IJCNLP)}, pages 4537--4546, Hong Kong,
  China, November. Association for Computational Linguistics.

\bibitem[\protect\citename{Dinan \bgroup et al.\egroup
  }2019b]{DBLP:conf/iclr/DinanRSFAW19}
Emily Dinan, Stephen Roller, Kurt Shuster, Angela Fan, Michael Auli, and Jason
  Weston.
\newblock 2019b.
\newblock Wizard of wikipedia: Knowledge-powered conversational agents.
\newblock In {\em 7th International Conference on Learning Representations,
  {ICLR} 2019, New Orleans, LA, USA, May 6-9, 2019}. OpenReview.net.

\bibitem[\protect\citename{Dinan \bgroup et al.\egroup }2022]{safetykit}
Emily Dinan, Gavin Abercrombie, A.~Bergman, Shannon Spruit, Dirk Hovy, Y-Lan
  Boureau, and Verena Rieser.
\newblock 2022.
\newblock {S}afety{K}it: First aid for measuring safety in open-domain
  conversational systems.
\newblock In {\em Proceedings of the 60th Annual Meeting of the Association for
  Computational Linguistics (Volume 1: Long Papers)}, pages 4113--4133, Dublin,
  Ireland, May. Association for Computational Linguistics.

\bibitem[\protect\citename{Dong \bgroup et al.\egroup
  }2023]{DBLP:journals/corr/abs-2301-00234}
Qingxiu Dong, Lei Li, Damai Dai, Ce~Zheng, Zhiyong Wu, Baobao Chang, Xu~Sun,
  Jingjing Xu, Lei Li, and Zhifang Sui.
\newblock 2023.
\newblock A survey for in-context learning.
\newblock {\em CoRR}, abs/2301.00234.

\bibitem[\protect\citename{Du \bgroup et al.\egroup
  }2022]{DBLP:conf/acl/DuDX0022}
Li~Du, Xiao Ding, Kai Xiong, Ting Liu, and Bing Qin.
\newblock 2022.
\newblock e-care: a new dataset for exploring explainable causal reasoning.
\newblock In Smaranda Muresan, Preslav Nakov, and Aline Villavicencio, editors,
  {\em Proceedings of the 60th Annual Meeting of the Association for
  Computational Linguistics (Volume 1: Long Papers), {ACL} 2022, Dublin,
  Ireland, May 22-27, 2022}, pages 432--446. Association for Computational
  Linguistics.

\bibitem[\protect\citename{D{\"u}ndar-Coecke}2022]{DndarCoecke2022ToWE}
Selma D{\"u}ndar-Coecke.
\newblock 2022.
\newblock To what extent is general intelligence relevant to causal reasoning?
  a developmental study.
\newblock {\em Frontiers in Psychology}, 13.

\bibitem[\protect\citename{Eric \bgroup et al.\egroup
  }2017]{DBLP:conf/sigdial/EricKCM17}
Mihail Eric, Lakshmi Krishnan, Fran{\c{c}}ois Charette, and Christopher~D.
  Manning.
\newblock 2017.
\newblock Key-value retrieval networks for task-oriented dialogue.
\newblock In Kristiina Jokinen, Manfred Stede, David DeVault, and Annie Louis,
  editors, {\em Proceedings of the 18th Annual SIGdial Meeting on Discourse and
  Dialogue, Saarbr{\"{u}}cken, Germany, August 15-17, 2017}, pages 37--49.
  Association for Computational Linguistics.

\bibitem[\protect\citename{Evans \bgroup et al.\egroup
  }2021]{DBLP:journals/corr/abs-2110-06674}
Owain Evans, Owen Cotton{-}Barratt, Lukas Finnveden, Adam Bales, Avital Balwit,
  Peter Wills, Luca Righetti, and William Saunders.
\newblock 2021.
\newblock Truthful {AI:} developing and governing {AI} that does not lie.
\newblock {\em CoRR}, abs/2110.06674.

\bibitem[\protect\citename{Evans}2002]{Evans2002LogicAH}
Jonathan Evans.
\newblock 2002.
\newblock Logic and human reasoning: an assessment of the deduction paradigm.
\newblock {\em Psychological bulletin}, 128 6:978--96.

\bibitem[\protect\citename{Forbes \bgroup et al.\egroup
  }2020]{SocialChemistry101}
Maxwell Forbes, Jena~D. Hwang, Vered Shwartz, Maarten Sap, and Yejin Choi.
\newblock 2020.
\newblock Social chemistry 101: Learning to reason about social and moral
  norms.
\newblock In {\em Proceedings of the 2020 Conference on Empirical Methods in
  Natural Language Processing (EMNLP)}, pages 653--670, Online, November.
  Association for Computational Linguistics.

\bibitem[\protect\citename{Fromkin \bgroup et al.\egroup
  }2018]{fromkin2018introduction}
Victoria Fromkin, Robert Rodman, and Nina Hyams.
\newblock 2018.
\newblock {\em An Introduction to Language (w/MLA9E Updates)}.
\newblock Cengage Learning.

\bibitem[\protect\citename{Gao \bgroup et al.\egroup }2021]{eval-harness}
Leo Gao, Jonathan Tow, Stella Biderman, Sid Black, Anthony DiPofi, Charles
  Foster, Laurence Golding, Jeffrey Hsu, Kyle McDonell, Niklas Muennighoff,
  Jason Phang, Laria Reynolds, Eric Tang, Anish Thite, Ben Wang, Kevin Wang,
  and Andy Zou.
\newblock 2021.
\newblock A framework for few-shot language model evaluation, September.

\bibitem[\protect\citename{Gao \bgroup et al.\egroup
  }2023]{DBLP:journals/corr/abs-2305-07375}
Jinglong Gao, Xiao Ding, Bing Qin, and Ting Liu.
\newblock 2023.
\newblock Is chatgpt a good causal reasoner? {A} comprehensive evaluation.
\newblock {\em CoRR}, abs/2305.07375.

\bibitem[\protect\citename{Gehman \bgroup et al.\egroup
  }2020]{realtoxicityprompts}
Samuel Gehman, Suchin Gururangan, Maarten Sap, Yejin Choi, and Noah~A. Smith.
\newblock 2020.
\newblock {R}eal{T}oxicity{P}rompts: Evaluating neural toxic degeneration in
  language models.
\newblock In {\em Findings of the Association for Computational Linguistics:
  EMNLP 2020}, pages 3356--3369, Online, November. Association for
  Computational Linguistics.

\bibitem[\protect\citename{Geva \bgroup et al.\egroup
  }2021]{DBLP:journals/tacl/GevaKSKRB21}
Mor Geva, Daniel Khashabi, Elad Segal, Tushar Khot, Dan Roth, and Jonathan
  Berant.
\newblock 2021.
\newblock Did aristotle use a laptop? {A} question answering benchmark with
  implicit reasoning strategies.
\newblock {\em Trans. Assoc. Comput. Linguistics}, 9:346--361.

\bibitem[\protect\citename{Gordon \bgroup et al.\egroup
  }2012]{DBLP:conf/SemEval/GordonKR12}
Andrew~S. Gordon, Zornitsa Kozareva, and Melissa Roemmele.
\newblock 2012.
\newblock Semeval-2012 task 7: Choice of plausible alternatives: An evaluation
  of commonsense causal reasoning.
\newblock In Eneko Agirre, Johan Bos, and Mona~T. Diab, editors, {\em
  Proceedings of the 6th International Workshop on Semantic Evaluation,
  SemEval@NAACL-HLT 2012, Montr{\'{e}}al, Canada, June 7-8, 2012}, pages
  394--398. The Association for Computer Linguistics.

\bibitem[\protect\citename{Hayati \bgroup et al.\egroup
  }2021]{DBLP:conf/emnlp/HayatiKU21}
Shirley~Anugrah Hayati, Dongyeop Kang, and Lyle Ungar.
\newblock 2021.
\newblock Does {BERT} learn as humans perceive? understanding linguistic styles
  through lexica.
\newblock In Marie{-}Francine Moens, Xuanjing Huang, Lucia Specia, and
  Scott~Wen{-}tau Yih, editors, {\em Proceedings of the 2021 Conference on
  Empirical Methods in Natural Language Processing, {EMNLP} 2021, Virtual Event
  / Punta Cana, Dominican Republic, 7-11 November, 2021}, pages 6323--6331.
  Association for Computational Linguistics.

\bibitem[\protect\citename{Hessel \bgroup et al.\egroup
  }2022]{DBLP:conf/eccv/HesselHPZBRSC22}
Jack Hessel, Jena~D. Hwang, Jae~Sung Park, Rowan Zellers, Chandra Bhagavatula,
  Anna Rohrbach, Kate Saenko, and Yejin Choi.
\newblock 2022.
\newblock The abduction of sherlock holmes: {A} dataset for visual abductive
  reasoning.
\newblock In Shai Avidan, Gabriel~J. Brostow, Moustapha Ciss{\'{e}},
  Giovanni~Maria Farinella, and Tal Hassner, editors, {\em Computer Vision -
  {ECCV} 2022 - 17th European Conference, Tel Aviv, Israel, October 23-27,
  2022, Proceedings, Part {XXXVI}}, volume 13696 of {\em Lecture Notes in
  Computer Science}, pages 558--575. Springer.

\bibitem[\protect\citename{Hoffmann}1999]{hoffmann1999meanings}
Terrence Hoffmann.
\newblock 1999.
\newblock The meanings of competency.
\newblock {\em Journal of european industrial training}, 23(6):275--286.

\bibitem[\protect\citename{Hosseini \bgroup et al.\egroup
  }2014]{DBLP:conf/emnlp/HosseiniHEK14}
Mohammad~Javad Hosseini, Hannaneh Hajishirzi, Oren Etzioni, and Nate Kushman.
\newblock 2014.
\newblock Learning to solve arithmetic word problems with verb categorization.
\newblock In Alessandro Moschitti, Bo~Pang, and Walter Daelemans, editors, {\em
  Proceedings of the 2014 Conference on Empirical Methods in Natural Language
  Processing, {EMNLP} 2014, October 25-29, 2014, Doha, Qatar, {A} meeting of
  SIGDAT, a Special Interest Group of the {ACL}}, pages 523--533. {ACL}.

\bibitem[\protect\citename{Huang \bgroup et al.\egroup
  }2022]{DBLP:conf/icml/HuangAPM22}
Wenlong Huang, Pieter Abbeel, Deepak Pathak, and Igor Mordatch.
\newblock 2022.
\newblock Language models as zero-shot planners: Extracting actionable
  knowledge for embodied agents.
\newblock In Kamalika Chaudhuri, Stefanie Jegelka, Le~Song, Csaba
  Szepesv{\'{a}}ri, Gang Niu, and Sivan Sabato, editors, {\em International
  Conference on Machine Learning, {ICML} 2022, 17-23 July 2022, Baltimore,
  Maryland, {USA}}, volume 162 of {\em Proceedings of Machine Learning
  Research}, pages 9118--9147. {PMLR}.

\bibitem[\protect\citename{Huang \bgroup et al.\egroup
  }2023]{DBLP:journals/corr/abs-2305-08322}
Yuzhen Huang, Yuzhuo Bai, Zhihao Zhu, Junlei Zhang, Jinghan Zhang, Tangjun Su,
  Junteng Liu, Chuancheng Lv, Yikai Zhang, Jiayi Lei, Yao Fu, Maosong Sun, and
  Junxian He.
\newblock 2023.
\newblock C-eval: {A} multi-level multi-discipline chinese evaluation suite for
  foundation models.
\newblock {\em CoRR}, abs/2305.08322.

\bibitem[\protect\citename{Hwang \bgroup et al.\egroup }2021]{Atomic}
Jena~D. Hwang, Chandra Bhagavatula, Ronan~Le Bras, Jeff Da, Keisuke Sakaguchi,
  Antoine Bosselut, and Yejin Choi.
\newblock 2021.
\newblock (comet-) atomic 2020: On symbolic and neural commonsense knowledge
  graphs.
\newblock In {\em Thirty-Fifth {AAAI} Conference on Artificial Intelligence,
  {AAAI} 2021, Thirty-Third Conference on Innovative Applications of Artificial
  Intelligence, {IAAI} 2021, The Eleventh Symposium on Educational Advances in
  Artificial Intelligence, {EAAI} 2021, Virtual Event, February 2-9, 2021},
  pages 6384--6392. {AAAI} Press.

\bibitem[\protect\citename{Iyyer \bgroup et al.\egroup
  }2017]{DBLP:conf/acl/IyyerYC17}
Mohit Iyyer, Wen{-}tau Yih, and Ming{-}Wei Chang.
\newblock 2017.
\newblock Search-based neural structured learning for sequential question
  answering.
\newblock In Regina Barzilay and Min{-}Yen Kan, editors, {\em Proceedings of
  the 55th Annual Meeting of the Association for Computational Linguistics,
  {ACL} 2017, Vancouver, Canada, July 30 - August 4, Volume 1: Long Papers},
  pages 1821--1831. Association for Computational Linguistics.

\bibitem[\protect\citename{Ji \bgroup et al.\egroup
  }2023]{DBLP:journals/csur/JiLFYSXIBMF23}
Ziwei Ji, Nayeon Lee, Rita Frieske, Tiezheng Yu, Dan Su, Yan Xu, Etsuko Ishii,
  Yejin Bang, Andrea Madotto, and Pascale Fung.
\newblock 2023.
\newblock Survey of hallucination in natural language generation.
\newblock {\em {ACM} Comput. Surv.}, 55(12):248:1--248:38.

\bibitem[\protect\citename{Jiang \bgroup et al.\egroup
  }2023]{DBLP:journals/corr/abs-2305-09645}
Jinhao Jiang, Kun Zhou, Zican Dong, Keming Ye, Wayne~Xin Zhao, and Ji{-}Rong
  Wen.
\newblock 2023.
\newblock Structgpt: {A} general framework for large language model to reason
  over structured data.
\newblock {\em CoRR}, abs/2305.09645.

\bibitem[\protect\citename{Joshi \bgroup et al.\egroup
  }2017]{DBLP:conf/acl/JoshiCWZ17}
Mandar Joshi, Eunsol Choi, Daniel~S. Weld, and Luke Zettlemoyer.
\newblock 2017.
\newblock Triviaqa: {A} large scale distantly supervised challenge dataset for
  reading comprehension.
\newblock In Regina Barzilay and Min{-}Yen Kan, editors, {\em Proceedings of
  the 55th Annual Meeting of the Association for Computational Linguistics,
  {ACL} 2017, Vancouver, Canada, July 30 - August 4, Volume 1: Long Papers},
  pages 1601--1611. Association for Computational Linguistics.

\bibitem[\protect\citename{Kadavath \bgroup et al.\egroup
  }2022]{DBLP:journals/corr/abs-2207-05221}
Saurav Kadavath, Tom Conerly, Amanda Askell, Tom Henighan, Dawn Drain, Ethan
  Perez, Nicholas Schiefer, Zac Hatfield{-}Dodds, Nova DasSarma, Eli
  Tran{-}Johnson, Scott Johnston, Sheer~El Showk, Andy Jones, Nelson Elhage,
  Tristan Hume, Anna Chen, Yuntao Bai, Sam Bowman, Stanislav Fort, Deep
  Ganguli, Danny Hernandez, Josh Jacobson, Jackson Kernion, Shauna Kravec,
  Liane Lovitt, Kamal Ndousse, Catherine Olsson, Sam Ringer, Dario Amodei, Tom
  Brown, Jack Clark, Nicholas Joseph, Ben Mann, Sam McCandlish, Chris Olah, and
  Jared Kaplan.
\newblock 2022.
\newblock Language models (mostly) know what they know.
\newblock {\em CoRR}, abs/2207.05221.

\bibitem[\protect\citename{Kakas and
  Michael}2020]{DBLP:journals/corr/abs-2010-12896}
Antonis~C. Kakas and Loizos Michael.
\newblock 2020.
\newblock Abduction and argumentation for explainable machine learning: {A}
  position survey.
\newblock {\em CoRR}, abs/2010.12896.

\bibitem[\protect\citename{Kaplan \bgroup et al.\egroup
  }2020]{DBLP:journals/corr/abs-2001-08361}
Jared Kaplan, Sam McCandlish, Tom Henighan, Tom~B. Brown, Benjamin Chess, Rewon
  Child, Scott Gray, Alec Radford, Jeffrey Wu, and Dario Amodei.
\newblock 2020.
\newblock Scaling laws for neural language models.
\newblock {\em CoRR}, abs/2001.08361.

\bibitem[\protect\citename{Kocoń \bgroup et al.\egroup
  }2021]{Offensiveaggressive}
Jan Kocoń, Alicja Figas, Marcin Gruza, Daria Puchalska, Tomasz Kajdanowicz,
  and Przemysław Kazienko.
\newblock 2021.
\newblock Offensive, aggressive, and hate speech analysis: From data-centric to
  human-centered approach.
\newblock {\em Information Processing \& Management}, 58(5):102643.

\bibitem[\protect\citename{Kojima \bgroup et al.\egroup
  }2022]{DBLP:conf/nips/KojimaGRMI22}
Takeshi Kojima, Shixiang~Shane Gu, Machel Reid, Yutaka Matsuo, and Yusuke
  Iwasawa.
\newblock 2022.
\newblock Large language models are zero-shot reasoners.
\newblock In {\em NeurIPS}.

\bibitem[\protect\citename{Krause \bgroup et al.\egroup }2021]{gedi}
Ben Krause, Akhilesh~Deepak Gotmare, Bryan McCann, Nitish~Shirish Keskar,
  Shafiq Joty, Richard Socher, and Nazneen~Fatema Rajani.
\newblock 2021.
\newblock {G}e{D}i: Generative discriminator guided sequence generation.
\newblock In {\em Findings of the Association for Computational Linguistics:
  EMNLP 2021}, pages 4929--4952, Punta Cana, Dominican Republic, November.
  Association for Computational Linguistics.

\bibitem[\protect\citename{K{\"{u}}{\c{c}}{\"{u}}k and Can}2021]{stance}
Dilek K{\"{u}}{\c{c}}{\"{u}}k and Fazli Can.
\newblock 2021.
\newblock Stance detection: {A} survey.
\newblock {\em {ACM} Comput. Surv.}, 53(1):12:1--12:37.

\bibitem[\protect\citename{Kuhn \bgroup et al.\egroup
  }2023]{DBLP:conf/iclr/KuhnGF23}
Lorenz Kuhn, Yarin Gal, and Sebastian Farquhar.
\newblock 2023.
\newblock Semantic uncertainty: Linguistic invariances for uncertainty
  estimation in natural language generation.
\newblock In {\em The Eleventh International Conference on Learning
  Representations, {ICLR} 2023, Kigali, Rwanda, May 1-5, 2023}. OpenReview.net.

\bibitem[\protect\citename{Lee \bgroup et al.\egroup
  }2018]{Lee2018HallucinationsIN}
Katherine Lee, Orhan Firat, Ashish Agarwal, Clara Fannjiang, and David
  Sussillo.
\newblock 2018.
\newblock Hallucinations in neural machine translation.

\bibitem[\protect\citename{Li \bgroup et al.\egroup }2022]{YouDontKnowMy}
Haoran Li, Yangqiu Song, and Lixin Fan.
\newblock 2022.
\newblock You don{'}t know my favorite color: Preventing dialogue
  representations from revealing speakers{'} private personas.
\newblock In {\em Proceedings of the 2022 Conference of the North American
  Chapter of the Association for Computational Linguistics: Human Language
  Technologies}, pages 5858--5870, Seattle, United States, July. Association
  for Computational Linguistics.

\bibitem[\protect\citename{Li \bgroup et al.\egroup
  }2023a]{DBLP:journals/corr/abs-2306-09212}
Haonan Li, Yixuan Zhang, Fajri Koto, Yifei Yang, Hai Zhao, Yeyun Gong, Nan
  Duan, and Timothy Baldwin.
\newblock 2023a.
\newblock {CMMLU:} measuring massive multitask language understanding in
  chinese.
\newblock {\em CoRR}, abs/2306.09212.

\bibitem[\protect\citename{Li \bgroup et al.\egroup
  }2023b]{DBLP:journals/corr/abs-2305-13269}
Xingxuan Li, Ruochen Zhao, Yew~Ken Chia, Bosheng Ding, Lidong Bing, Shafiq~R.
  Joty, and Soujanya Poria.
\newblock 2023b.
\newblock Chain of knowledge: {A} framework for grounding large language models
  with structured knowledge bases.
\newblock {\em CoRR}, abs/2305.13269.

\bibitem[\protect\citename{Liang \bgroup et al.\egroup
  }2022]{DBLP:journals/corr/abs-2211-09110}
Percy Liang, Rishi Bommasani, Tony Lee, Dimitris Tsipras, Dilara Soylu,
  Michihiro Yasunaga, Yian Zhang, Deepak Narayanan, Yuhuai Wu, Ananya Kumar,
  Benjamin Newman, Binhang Yuan, Bobby Yan, Ce~Zhang, Christian Cosgrove,
  Christopher~D. Manning, Christopher R{\'{e}}, Diana Acosta{-}Navas, Drew~A.
  Hudson, Eric Zelikman, Esin Durmus, Faisal Ladhak, Frieda Rong, Hongyu Ren,
  Huaxiu Yao, Jue Wang, Keshav Santhanam, Laurel~J. Orr, Lucia Zheng, Mert
  Y{\"{u}}ksekg{\"{o}}n{\"{u}}l, Mirac Suzgun, Nathan Kim, Neel Guha,
  Niladri~S. Chatterji, Omar Khattab, Peter Henderson, Qian Huang, Ryan Chi,
  Sang~Michael Xie, Shibani Santurkar, Surya Ganguli, Tatsunori Hashimoto,
  Thomas Icard, Tianyi Zhang, Vishrav Chaudhary, William Wang, Xuechen Li,
  Yifan Mai, Yuhui Zhang, and Yuta Koreeda.
\newblock 2022.
\newblock Holistic evaluation of language models.
\newblock {\em CoRR}, abs/2211.09110.

\bibitem[\protect\citename{Lin \bgroup et al.\egroup
  }2022a]{DBLP:journals/tmlr/LinHE22}
Stephanie Lin, Jacob Hilton, and Owain Evans.
\newblock 2022a.
\newblock Teaching models to express their uncertainty in words.
\newblock {\em Trans. Mach. Learn. Res.}, 2022.

\bibitem[\protect\citename{Lin \bgroup et al.\egroup
  }2022b]{DBLP:conf/acl/LinHE22}
Stephanie Lin, Jacob Hilton, and Owain Evans.
\newblock 2022b.
\newblock Truthfulqa: Measuring how models mimic human falsehoods.
\newblock In Smaranda Muresan, Preslav Nakov, and Aline Villavicencio, editors,
  {\em Proceedings of the 60th Annual Meeting of the Association for
  Computational Linguistics (Volume 1: Long Papers), {ACL} 2022, Dublin,
  Ireland, May 22-27, 2022}, pages 3214--3252. Association for Computational
  Linguistics.

\bibitem[\protect\citename{Lindstr{\"{o}}m and
  Abraham}2022]{DBLP:conf/nesy/LindstromA22}
Adam~Dahlgren Lindstr{\"{o}}m and Savitha~Sam Abraham.
\newblock 2022.
\newblock Clevr-math: {A} dataset for compositional language, visual and
  mathematical reasoning.
\newblock In Artur~S. d'Avila Garcez and Ernesto Jim{\'{e}}nez{-}Ruiz, editors,
  {\em Proceedings of the 16th International Workshop on Neural-Symbolic
  Learning and Reasoning as part of the 2nd International Joint Conference on
  Learning {\&} Reasoning {(IJCLR} 2022), Cumberland Lodge, Windsor Great Park,
  UK, September 28-30, 2022}, volume 3212 of {\em {CEUR} Workshop Proceedings},
  pages 155--170. CEUR-WS.org.

\bibitem[\protect\citename{Liu \bgroup et al.\egroup }2021]{ESC}
Siyang Liu, Chujie Zheng, Orianna Demasi, Sahand Sabour, Yu~Li, Zhou Yu, Yong
  Jiang, and Minlie Huang.
\newblock 2021.
\newblock Towards emotional support dialog systems.
\newblock In Chengqing Zong, Fei Xia, Wenjie Li, and Roberto Navigli, editors,
  {\em Proceedings of the 59th Annual Meeting of the Association for
  Computational Linguistics and the 11th International Joint Conference on
  Natural Language Processing, {ACL/IJCNLP} 2021, (Volume 1: Long Papers),
  Virtual Event, August 1-6, 2021}, pages 3469--3483. Association for
  Computational Linguistics.

\bibitem[\protect\citename{Liu \bgroup et al.\egroup
  }2023]{DBLP:journals/corr/abs-2305-01210}
Jiawei Liu, Chunqiu~Steven Xia, Yuyao Wang, and Lingming Zhang.
\newblock 2023.
\newblock Is your code generated by chatgpt really correct? rigorous evaluation
  of large language models for code generation.
\newblock {\em CoRR}, abs/2305.01210.

\bibitem[\protect\citename{Liu}2015]{SC_2015}
Bing Liu.
\newblock 2015.
\newblock {\em Sentiment Analysis - Mining Opinions, Sentiments, and Emotions}.
\newblock Cambridge University Press.

\bibitem[\protect\citename{Lu \bgroup et al.\egroup
  }2023]{DBLP:conf/iclr/Lu0CWZRCK23}
Pan Lu, Liang Qiu, Kai{-}Wei Chang, Ying~Nian Wu, Song{-}Chun Zhu, Tanmay
  Rajpurohit, Peter Clark, and Ashwin Kalyan.
\newblock 2023.
\newblock Dynamic prompt learning via policy gradient for semi-structured
  mathematical reasoning.
\newblock In {\em The Eleventh International Conference on Learning
  Representations, {ICLR} 2023, Kigali, Rwanda, May 1-5, 2023}. OpenReview.net.

\bibitem[\protect\citename{Lwowski \bgroup et al.\egroup
  }2022]{measuringgeopraphic}
Brandon Lwowski, Paul Rad, and Anthony Rios.
\newblock 2022.
\newblock Measuring geographic performance disparities of offensive language
  classifiers.
\newblock In {\em Proceedings of the 29th International Conference on
  Computational Linguistics}, pages 6600--6616, Gyeongju, Republic of Korea,
  October. International Committee on Computational Linguistics.

\bibitem[\protect\citename{Lyu \bgroup et al.\egroup
  }2023]{DBLP:journals/corr/abs-2301-13379}
Qing Lyu, Shreya Havaldar, Adam Stein, Li~Zhang, Delip Rao, Eric Wong, Marianna
  Apidianaki, and Chris Callison{-}Burch.
\newblock 2023.
\newblock Faithful chain-of-thought reasoning.
\newblock {\em CoRR}, abs/2301.13379.

\bibitem[\protect\citename{Maas \bgroup et al.\egroup }2011]{dataset_imdb}
Andrew~L. Maas, Raymond~E. Daly, Peter~T. Pham, Dan Huang, Andrew~Y. Ng, and
  Christopher Potts.
\newblock 2011.
\newblock Learning word vectors for sentiment analysis.
\newblock In Dekang Lin, Yuji Matsumoto, and Rada Mihalcea, editors, {\em The
  49th Annual Meeting of the Association for Computational Linguistics: Human
  Language Technologies, Proceedings of the Conference, 19-24 June, 2011,
  Portland, Oregon, {USA}}, pages 142--150. The Association for Computer
  Linguistics.

\bibitem[\protect\citename{Mankowitz \bgroup et al.\egroup
  }2023]{Mankowitz2023FasterSA}
Daniel~Jaymin Mankowitz, Andrea Michi, Anton Zhernov, Marco Gelmi, Marco Selvi,
  Cosmin Paduraru, Edouard Leurent, Shariq Iqbal, Jean-Baptiste Lespiau, Alex
  Ahern, Thomas K{\"o}ppe, Kevin Millikin, Stephen Gaffney, Sophie Elster,
  Jackson Broshear, Chris Gamble, Kieran Milan, Robert Tung, Minjae Hwang,
  taylan. cemgil, Mohammadamin Barekatain, Yujia Li, Amol Mandhane, Thomas
  Hubert, Julian Schrittwieser, Demis Hassabis, Pushmeet Kohli, Martin~A.
  Riedmiller, Oriol Vinyals, and David Silver.
\newblock 2023.
\newblock Faster sorting algorithms discovered using deep reinforcement
  learning.
\newblock {\em Nature}, 618:257 -- 263.

\bibitem[\protect\citename{Mathew \bgroup et al.\egroup
  }2021]{DBLP:conf/aaai/MathewSYBG021}
Binny Mathew, Punyajoy Saha, Seid~Muhie Yimam, Chris Biemann, Pawan Goyal, and
  Animesh Mukherjee.
\newblock 2021.
\newblock Hatexplain: {A} benchmark dataset for explainable hate speech
  detection.
\newblock In {\em Thirty-Fifth {AAAI} Conference on Artificial Intelligence,
  {AAAI} 2021, Thirty-Third Conference on Innovative Applications of Artificial
  Intelligence, {IAAI} 2021, The Eleventh Symposium on Educational Advances in
  Artificial Intelligence, {EAAI} 2021, Virtual Event, February 2-9, 2021},
  pages 14867--14875. {AAAI} Press.

\bibitem[\protect\citename{Mialon \bgroup et al.\egroup
  }2023]{DBLP:journals/corr/abs-2302-07842}
Gr{\'{e}}goire Mialon, Roberto Dess{\`{\i}}, Maria Lomeli, Christoforos
  Nalmpantis, Ramakanth Pasunuru, Roberta Raileanu, Baptiste Rozi{\`{e}}re,
  Timo Schick, Jane Dwivedi{-}Yu, Asli Celikyilmaz, Edouard Grave, Yann LeCun,
  and Thomas Scialom.
\newblock 2023.
\newblock Augmented language models: a survey.
\newblock {\em CoRR}, abs/2302.07842.

\bibitem[\protect\citename{Mielke \bgroup et al.\egroup
  }2022]{DBLP:journals/tacl/MielkeSDB22}
Sabrina~J. Mielke, Arthur Szlam, Emily Dinan, and Y{-}Lan Boureau.
\newblock 2022.
\newblock Reducing conversational agents' overconfidence through linguistic
  calibration.
\newblock {\em Trans. Assoc. Comput. Linguistics}, 10:857--872.

\bibitem[\protect\citename{Mihaylov \bgroup et al.\egroup
  }2018]{DBLP:conf/emnlp/MihaylovCKS18}
Todor Mihaylov, Peter Clark, Tushar Khot, and Ashish Sabharwal.
\newblock 2018.
\newblock Can a suit of armor conduct electricity? {A} new dataset for open
  book question answering.
\newblock In Ellen Riloff, David Chiang, Julia Hockenmaier, and Jun'ichi
  Tsujii, editors, {\em Proceedings of the 2018 Conference on Empirical Methods
  in Natural Language Processing, Brussels, Belgium, October 31 - November 4,
  2018}, pages 2381--2391. Association for Computational Linguistics.

\bibitem[\protect\citename{Mirza \bgroup et al.\egroup
  }2014]{Mirza2014AnnotatingCI}
Paramita Mirza, R.~Sprugnoli, Sara Tonelli, and Manuela Speranza.
\newblock 2014.
\newblock Annotating causality in the tempeval-3 corpus.
\newblock In {\em Conference of the European Chapter of the Association for
  Computational Linguistics}.

\bibitem[\protect\citename{Mishra \bgroup et al.\egroup
  }2022a]{DBLP:conf/emnlp/MishraFLTWBRTSC22}
Swaroop Mishra, Matthew Finlayson, Pan Lu, Leonard Tang, Sean Welleck, Chitta
  Baral, Tanmay Rajpurohit, Oyvind Tafjord, Ashish Sabharwal, Peter Clark, and
  Ashwin Kalyan.
\newblock 2022a.
\newblock {LILA:} {A} unified benchmark for mathematical reasoning.
\newblock In Yoav Goldberg, Zornitsa Kozareva, and Yue Zhang, editors, {\em
  Proceedings of the 2022 Conference on Empirical Methods in Natural Language
  Processing, {EMNLP} 2022, Abu Dhabi, United Arab Emirates, December 7-11,
  2022}, pages 5807--5832. Association for Computational Linguistics.

\bibitem[\protect\citename{Mishra \bgroup et al.\egroup
  }2022b]{DBLP:conf/acl/MishraMVSCBK22}
Swaroop Mishra, Arindam Mitra, Neeraj Varshney, Bhavdeep~Singh Sachdeva, Peter
  Clark, Chitta Baral, and Ashwin Kalyan.
\newblock 2022b.
\newblock Numglue: {A} suite of fundamental yet challenging mathematical
  reasoning tasks.
\newblock In Smaranda Muresan, Preslav Nakov, and Aline Villavicencio, editors,
  {\em Proceedings of the 60th Annual Meeting of the Association for
  Computational Linguistics (Volume 1: Long Papers), {ACL} 2022, Dublin,
  Ireland, May 22-27, 2022}, pages 3505--3523. Association for Computational
  Linguistics.

\bibitem[\protect\citename{Nadeem \bgroup et al.\egroup }2021]{stereoset}
Moin Nadeem, Anna Bethke, and Siva Reddy.
\newblock 2021.
\newblock {S}tereo{S}et: Measuring stereotypical bias in pretrained language
  models.
\newblock In {\em Proceedings of the 59th Annual Meeting of the Association for
  Computational Linguistics and the 11th International Joint Conference on
  Natural Language Processing (Volume 1: Long Papers)}, pages 5356--5371,
  Online, August. Association for Computational Linguistics.

\bibitem[\protect\citename{Nan \bgroup et al.\egroup }2021]{Nan2021FeTaQAFT}
Linyong Nan, Chia-Hsuan Hsieh, Ziming Mao, Xi~Victoria Lin, Neha Verma, Rui
  Zhang, Wojciech Kryscinski, Nick Schoelkopf, Riley Kong, Xiangru Tang, Murori
  Mutuma, Benjamin Rosand, Isabel Trindade, Renusree Bandaru, Jacob Cunningham,
  Caiming Xiong, and Dragomir~R. Radev.
\newblock 2021.
\newblock Fetaqa: Free-form table question answering.
\newblock {\em Transactions of the Association for Computational Linguistics},
  10:35--49.

\bibitem[\protect\citename{Nangia \bgroup et al.\egroup }2020]{crowspairs}
Nikita Nangia, Clara Vania, Rasika Bhalerao, and Samuel~R. Bowman.
\newblock 2020.
\newblock {C}row{S}-pairs: A challenge dataset for measuring social biases in
  masked language models.
\newblock In {\em Proceedings of the 2020 Conference on Empirical Methods in
  Natural Language Processing (EMNLP)}, pages 1953--1967, Online, November.
  Association for Computational Linguistics.

\bibitem[\protect\citename{Olsson \bgroup et al.\egroup
  }2022]{DBLP:journals/corr/abs-2209-11895}
Catherine Olsson, Nelson Elhage, Neel Nanda, Nicholas Joseph, Nova DasSarma,
  Tom Henighan, Ben Mann, Amanda Askell, Yuntao Bai, Anna Chen, Tom Conerly,
  Dawn Drain, Deep Ganguli, Zac Hatfield{-}Dodds, Danny Hernandez, Scott
  Johnston, Andy Jones, Jackson Kernion, Liane Lovitt, Kamal Ndousse, Dario
  Amodei, Tom Brown, Jack Clark, Jared Kaplan, Sam McCandlish, and Chris Olah.
\newblock 2022.
\newblock In-context learning and induction heads.
\newblock {\em CoRR}, abs/2209.11895.

\bibitem[\protect\citename{OpenAI}2023]{DBLP:journals/corr/abs-2303-08774}
OpenAI.
\newblock 2023.
\newblock {GPT-4} technical report.
\newblock {\em CoRR}, abs/2303.08774.

\bibitem[\protect\citename{Ouyang \bgroup et al.\egroup
  }2022]{DBLP:conf/nips/Ouyang0JAWMZASR22}
Long Ouyang, Jeffrey Wu, Xu~Jiang, Diogo Almeida, Carroll~L. Wainwright, Pamela
  Mishkin, Chong Zhang, Sandhini Agarwal, Katarina Slama, Alex Ray, John
  Schulman, Jacob Hilton, Fraser Kelton, Luke Miller, Maddie Simens, Amanda
  Askell, Peter Welinder, Paul~F. Christiano, Jan Leike, and Ryan Lowe.
\newblock 2022.
\newblock Training language models to follow instructions with human feedback.
\newblock In {\em NeurIPS}.

\bibitem[\protect\citename{Ovchinnikova}2012]{DBLP:books/daglib/0028671}
Ekaterina Ovchinnikova.
\newblock 2012.
\newblock {\em Integration of World Knowledge for Natural Language
  Understanding}, volume~3 of {\em Atlantis Thinking Machines}.
\newblock Atlantis Press.

\bibitem[\protect\citename{Pallagani \bgroup et al.\egroup
  }2023]{DBLP:journals/corr/abs-2305-16151}
Vishal Pallagani, Bharath Muppasani, Keerthiram Murugesan, Francesca Rossi,
  Biplav Srivastava, Lior Horesh, Francesco Fabiano, and Andrea Loreggia.
\newblock 2023.
\newblock Understanding the capabilities of large language models for automated
  planning.
\newblock {\em CoRR}, abs/2305.16151.

\bibitem[\protect\citename{Papineni \bgroup et al.\egroup }2002]{bleu}
Kishore Papineni, Salim Roukos, Todd Ward, and Wei{-}Jing Zhu.
\newblock 2002.
\newblock Bleu: a method for automatic evaluation of machine translation.
\newblock In {\em Proceedings of the 40th Annual Meeting of the Association for
  Computational Linguistics, July 6-12, 2002, Philadelphia, PA, {USA}}, pages
  311--318. {ACL}.

\bibitem[\protect\citename{Parikh \bgroup et al.\egroup
  }2020]{DBLP:conf/emnlp/ParikhWGFDYD20}
Ankur~P. Parikh, Xuezhi Wang, Sebastian Gehrmann, Manaal Faruqui, Bhuwan
  Dhingra, Diyi Yang, and Dipanjan Das.
\newblock 2020.
\newblock Totto: {A} controlled table-to-text generation dataset.
\newblock In Bonnie Webber, Trevor Cohn, Yulan He, and Yang Liu, editors, {\em
  Proceedings of the 2020 Conference on Empirical Methods in Natural Language
  Processing, {EMNLP} 2020, Online, November 16-20, 2020}, pages 1173--1186.
  Association for Computational Linguistics.

\bibitem[\protect\citename{Pasupat and Liang}2015]{DBLP:conf/acl/PasupatL15}
Panupong Pasupat and Percy Liang.
\newblock 2015.
\newblock Compositional semantic parsing on semi-structured tables.
\newblock In {\em Proceedings of the 53rd Annual Meeting of the Association for
  Computational Linguistics and the 7th International Joint Conference on
  Natural Language Processing of the Asian Federation of Natural Language
  Processing, {ACL} 2015, July 26-31, 2015, Beijing, China, Volume 1: Long
  Papers}, pages 1470--1480. The Association for Computer Linguistics.

\bibitem[\protect\citename{Ponti \bgroup et al.\egroup
  }2020]{DBLP:conf/emnlp/PontiGMLVK20}
Edoardo~Maria Ponti, Goran Glavas, Olga Majewska, Qianchu Liu, Ivan Vulic, and
  Anna Korhonen.
\newblock 2020.
\newblock {XCOPA:} {A} multilingual dataset for causal commonsense reasoning.
\newblock In Bonnie Webber, Trevor Cohn, Yulan He, and Yang Liu, editors, {\em
  Proceedings of the 2020 Conference on Empirical Methods in Natural Language
  Processing, {EMNLP} 2020, Online, November 16-20, 2020}, pages 2362--2376.
  Association for Computational Linguistics.

\bibitem[\protect\citename{Pontiki \bgroup et al.\egroup
  }2014]{SemEval-2014_Task4}
Maria Pontiki, Dimitris Galanis, John Pavlopoulos, Harris Papageorgiou, Ion
  Androutsopoulos, and Suresh Manandhar.
\newblock 2014.
\newblock Semeval-2014 task 4: Aspect based sentiment analysis.
\newblock In Preslav Nakov and Torsten Zesch, editors, {\em Proceedings of the
  8th International Workshop on Semantic Evaluation, SemEval@COLING 2014,
  Dublin, Ireland, August 23-24, 2014}, pages 27--35. The Association for
  Computer Linguistics.

\bibitem[\protect\citename{Pontiki \bgroup et al.\egroup
  }2015]{SemEval-2015_Task12}
Maria Pontiki, Dimitris Galanis, Haris Papageorgiou, Suresh Manandhar, and Ion
  Androutsopoulos.
\newblock 2015.
\newblock Semeval-2015 task 12: Aspect based sentiment analysis.
\newblock In Daniel~M. Cer, David Jurgens, Preslav Nakov, and Torsten Zesch,
  editors, {\em Proceedings of the 9th International Workshop on Semantic
  Evaluation, SemEval@NAACL-HLT 2015, Denver, Colorado, USA, June 4-5, 2015},
  pages 486--495. The Association for Computer Linguistics.

\bibitem[\protect\citename{Pontiki \bgroup et al.\egroup
  }2016]{semeval2016_Task5}
Maria Pontiki, Dimitris Galanis, Haris Papageorgiou, Ion Androutsopoulos,
  Suresh Manandhar, Mohammad AL-Smadi, Mahmoud Al-Ayyoub, Yanyan Zhao, Bing
  Qin, Orph{\'e}e De~Clercq, V{\'e}ronique Hoste, Marianna Apidianaki, Xavier
  Tannier, Natalia Loukachevitch, Evgeniy Kotelnikov, Nuria Bel,
  Salud~Mar{\'\i}a Jim{\'e}nez-Zafra, and G{\"u}l{\c{s}}en Eryi{\u{g}}it.
\newblock 2016.
\newblock {S}em{E}val-2016 task 5: Aspect based sentiment analysis.
\newblock In {\em Proceedings of the 10th International Workshop on Semantic
  Evaluation ({S}em{E}val-2016)}, pages 19--30, San Diego, California, June.
  Association for Computational Linguistics.

\bibitem[\protect\citename{Poria \bgroup et al.\egroup }2023]{MAST-ref}
Soujanya Poria, Devamanyu Hazarika, Navonil Majumder, and Rada Mihalcea.
\newblock 2023.
\newblock Beneath the tip of the iceberg: Current challenges and new directions
  in sentiment analysis research.
\newblock {\em {IEEE} Trans. Affect. Comput.}, 14(1):108--132.

\bibitem[\protect\citename{Pradhan \bgroup et al.\egroup }2020]{offensive}
Rahul Pradhan, Ankur Chaturvedi, Aprna Tripathi, and Dilip~Kumar Sharma.
\newblock 2020.
\newblock A review on offensive language detection.
\newblock In Mohan~L. Kolhe, Shailesh Tiwari, Munesh~C. Trivedi, and Krishn~K.
  Mishra, editors, {\em Advances in Data and Information Sciences}, pages
  433--439, Singapore. Springer Singapore.

\bibitem[\protect\citename{Qiao \bgroup et al.\egroup
  }2022]{DBLP:journals/corr/abs-2212-09597}
Shuofei Qiao, Yixin Ou, Ningyu Zhang, Xiang Chen, Yunzhi Yao, Shumin Deng,
  Chuanqi Tan, Fei Huang, and Huajun Chen.
\newblock 2022.
\newblock Reasoning with language model prompting: {A} survey.
\newblock {\em CoRR}, abs/2212.09597.

\bibitem[\protect\citename{Qiu \bgroup et al.\egroup
  }2019]{DBLP:conf/acl/QiuXQZLZY19}
Lin Qiu, Yunxuan Xiao, Yanru Qu, Hao Zhou, Lei Li, Weinan Zhang, and Yong Yu.
\newblock 2019.
\newblock Dynamically fused graph network for multi-hop reasoning.
\newblock In Anna Korhonen, David~R. Traum, and Llu{\'{\i}}s M{\`{a}}rquez,
  editors, {\em Proceedings of the 57th Conference of the Association for
  Computational Linguistics, {ACL} 2019, Florence, Italy, July 28- August 2,
  2019, Volume 1: Long Papers}, pages 6140--6150. Association for Computational
  Linguistics.

\bibitem[\protect\citename{Rashkin \bgroup et al.\egroup }2019]{emp}
Hannah Rashkin, Eric~Michael Smith, Margaret Li, and Y{-}Lan Boureau.
\newblock 2019.
\newblock Towards empathetic open-domain conversation models: {A} new benchmark
  and dataset.
\newblock In Anna Korhonen, David~R. Traum, and Llu{\'{\i}}s M{\`{a}}rquez,
  editors, {\em Proceedings of the 57th Conference of the Association for
  Computational Linguistics, {ACL} 2019, Florence, Italy, July 28- August 2,
  2019, Volume 1: Long Papers}, pages 5370--5381. Association for Computational
  Linguistics.

\bibitem[\protect\citename{Rosenthal \bgroup et al.\egroup
  }2017]{dataset_twitter_SSC}
Sara Rosenthal, Noura Farra, and Preslav Nakov.
\newblock 2017.
\newblock Semeval-2017 task 4: Sentiment analysis in twitter.
\newblock In Steven Bethard, Marine Carpuat, Marianna Apidianaki, Saif~M.
  Mohammad, Daniel~M. Cer, and David Jurgens, editors, {\em Proceedings of the
  11th International Workshop on Semantic Evaluation, SemEval@ACL 2017,
  Vancouver, Canada, August 3-4, 2017}, pages 502--518. Association for
  Computational Linguistics.

\bibitem[\protect\citename{Sailunaz \bgroup et al.\egroup }2018]{emotion}
Kashfia Sailunaz, Manmeet Dhaliwal, Jon~G. Rokne, and Reda Alhajj.
\newblock 2018.
\newblock Emotion detection from text and speech: a survey.
\newblock {\em Soc. Netw. Anal. Min.}, 8(1):28:1--28:26.

\bibitem[\protect\citename{Sap \bgroup et al.\egroup }2020]{SocialBiasFrames}
Maarten Sap, Saadia Gabriel, Lianhui Qin, Dan Jurafsky, Noah~A. Smith, and
  Yejin Choi.
\newblock 2020.
\newblock Social bias frames: Reasoning about social and power implications of
  language.
\newblock In {\em Proceedings of the 58th Annual Meeting of the Association for
  Computational Linguistics}, pages 5477--5490, Online, July. Association for
  Computational Linguistics.

\bibitem[\protect\citename{Schmidt and Wiegand}2017]{hate}
Anna Schmidt and Michael Wiegand.
\newblock 2017.
\newblock A survey on hate speech detection using natural language processing.
\newblock In Lun{-}Wei Ku and Cheng{-}Te Li, editors, {\em Proceedings of the
  Fifth International Workshop on Natural Language Processing for Social Media,
  SocialNLP@EACL 2017, Valencia, Spain, April 3, 2017}, pages 1--10.
  Association for Computational Linguistics.

\bibitem[\protect\citename{Shi \bgroup et al.\egroup
  }2023]{DBLP:conf/iclr/ShiSF0SVCTRZ0W23}
Freda Shi, Mirac Suzgun, Markus Freitag, Xuezhi Wang, Suraj Srivats, Soroush
  Vosoughi, Hyung~Won Chung, Yi~Tay, Sebastian Ruder, Denny Zhou, Dipanjan Das,
  and Jason Wei.
\newblock 2023.
\newblock Language models are multilingual chain-of-thought reasoners.
\newblock In {\em The Eleventh International Conference on Learning
  Representations, {ICLR} 2023, Kigali, Rwanda, May 1-5, 2023}. OpenReview.net.

\bibitem[\protect\citename{Sinha \bgroup et al.\egroup
  }2019]{DBLP:conf/emnlp/SinhaSDPH19}
Koustuv Sinha, Shagun Sodhani, Jin Dong, Joelle Pineau, and William~L.
  Hamilton.
\newblock 2019.
\newblock {CLUTRR:} {A} diagnostic benchmark for inductive reasoning from text.
\newblock In Kentaro Inui, Jing Jiang, Vincent Ng, and Xiaojun Wan, editors,
  {\em Proceedings of the 2019 Conference on Empirical Methods in Natural
  Language Processing and the 9th International Joint Conference on Natural
  Language Processing, {EMNLP-IJCNLP} 2019, Hong Kong, China, November 3-7,
  2019}, pages 4505--4514. Association for Computational Linguistics.

\bibitem[\protect\citename{Socher \bgroup et al.\egroup }2013]{dataset_SST}
Richard Socher, Alex Perelygin, Jean Wu, Jason Chuang, Christopher~D. Manning,
  Andrew Ng, and Christopher Potts.
\newblock 2013.
\newblock Recursive deep models for semantic compositionality over a sentiment
  treebank.
\newblock In {\em Proceedings of the 2013 Conference on Empirical Methods in
  Natural Language Processing}, pages 1631--1642, Seattle, Washington, USA,
  October. Association for Computational Linguistics.

\bibitem[\protect\citename{Sodian and Kristen}2010]{tom}
Beate Sodian and Susanne Kristen, 2010.
\newblock {\em Theory of Mind}, pages 189--201.
\newblock Springer Berlin Heidelberg, Berlin, Heidelberg.

\bibitem[\protect\citename{Speer \bgroup et al.\egroup }2017]{conceptnet}
Robyn Speer, Joshua Chin, and Catherine Havasi.
\newblock 2017.
\newblock Conceptnet 5.5: An open multilingual graph of general knowledge.
\newblock In Satinder Singh and Shaul Markovitch, editors, {\em Proceedings of
  the Thirty-First {AAAI} Conference on Artificial Intelligence, February 4-9,
  2017, San Francisco, California, {USA}}, pages 4444--4451. {AAAI} Press.

\bibitem[\protect\citename{Sridhara \bgroup et al.\egroup
  }2023]{DBLP:journals/corr/abs-2305-16837}
Giriprasad Sridhara, Ranjani~H. G., and Sourav Mazumdar.
\newblock 2023.
\newblock Chatgpt: {A} study on its utility for ubiquitous software engineering
  tasks.
\newblock {\em CoRR}, abs/2305.16837.

\bibitem[\protect\citename{Srivastava \bgroup et al.\egroup
  }2022]{DBLP:journals/corr/abs-2206-04615}
Aarohi Srivastava, Abhinav Rastogi, Abhishek Rao, Abu Awal~Md Shoeb, Abubakar
  Abid, Adam Fisch, Adam~R. Brown, Adam Santoro, Aditya Gupta, Adri{\`{a}}
  Garriga{-}Alonso, Agnieszka Kluska, Aitor Lewkowycz, Akshat Agarwal, Alethea
  Power, Alex Ray, Alex Warstadt, Alexander~W. Kocurek, Ali Safaya, Ali Tazarv,
  Alice Xiang, Alicia Parrish, Allen Nie, Aman Hussain, Amanda Askell, Amanda
  Dsouza, Ameet Rahane, Anantharaman~S. Iyer, Anders Andreassen, Andrea
  Santilli, Andreas Stuhlm{\"{u}}ller, Andrew~M. Dai, Andrew La, Andrew~K.
  Lampinen, Andy Zou, Angela Jiang, Angelica Chen, Anh Vuong, Animesh Gupta,
  Anna Gottardi, Antonio Norelli, Anu Venkatesh, Arash Gholamidavoodi, Arfa
  Tabassum, Arul Menezes, Arun Kirubarajan, Asher Mullokandov, Ashish
  Sabharwal, Austin Herrick, Avia Efrat, Aykut Erdem, Ayla Karakas, and et~al.
\newblock 2022.
\newblock Beyond the imitation game: Quantifying and extrapolating the
  capabilities of language models.
\newblock {\em CoRR}, abs/2206.04615.

\bibitem[\protect\citename{Storks \bgroup et al.\egroup
  }2019]{DBLP:journals/corr/abs-1904-01172}
Shane Storks, Qiaozi Gao, and Joyce~Y. Chai.
\newblock 2019.
\newblock Commonsense reasoning for natural language understanding: {A} survey
  of benchmarks, resources, and approaches.
\newblock {\em CoRR}, abs/1904.01172.

\bibitem[\protect\citename{Sun \bgroup et al.\egroup }2022]{onthesafetyof}
Hao Sun, Guangxuan Xu, Jiawen Deng, Jiale Cheng, Chujie Zheng, Hao Zhou, Nanyun
  Peng, Xiaoyan Zhu, and Minlie Huang.
\newblock 2022.
\newblock On the safety of conversational models: Taxonomy, dataset, and
  benchmark.
\newblock In {\em Findings of the Association for Computational Linguistics:
  ACL 2022}, pages 3906--3923, Dublin, Ireland, May. Association for
  Computational Linguistics.

\bibitem[\protect\citename{Sur{\'{\i}}s \bgroup et al.\egroup
  }2023]{DBLP:journals/corr/abs-2303-08128}
D{\'{\i}}dac Sur{\'{\i}}s, Sachit Menon, and Carl Vondrick.
\newblock 2023.
\newblock Vipergpt: Visual inference via python execution for reasoning.
\newblock {\em CoRR}, abs/2303.08128.

\bibitem[\protect\citename{Szegedy \bgroup et al.\egroup
  }2014]{DBLP:journals/corr/SzegedyZSBEGF13}
Christian Szegedy, Wojciech Zaremba, Ilya Sutskever, Joan Bruna, Dumitru Erhan,
  Ian~J. Goodfellow, and Rob Fergus.
\newblock 2014.
\newblock Intriguing properties of neural networks.
\newblock In Yoshua Bengio and Yann LeCun, editors, {\em 2nd International
  Conference on Learning Representations, {ICLR} 2014, Banff, AB, Canada, April
  14-16, 2014, Conference Track Proceedings}.

\bibitem[\protect\citename{Talmor \bgroup et al.\egroup
  }2019]{DBLP:conf/naacl/TalmorHLB19}
Alon Talmor, Jonathan Herzig, Nicholas Lourie, and Jonathan Berant.
\newblock 2019.
\newblock Commonsenseqa: {A} question answering challenge targeting commonsense
  knowledge.
\newblock In Jill Burstein, Christy Doran, and Thamar Solorio, editors, {\em
  Proceedings of the 2019 Conference of the North American Chapter of the
  Association for Computational Linguistics: Human Language Technologies,
  {NAACL-HLT} 2019, Minneapolis, MN, USA, June 2-7, 2019, Volume 1 (Long and
  Short Papers)}, pages 4149--4158. Association for Computational Linguistics.

\bibitem[\protect\citename{Talmor \bgroup et al.\egroup
  }2021]{DBLP:conf/iclr/TalmorYCLWAIHB21}
Alon Talmor, Ori Yoran, Amnon Catav, Dan Lahav, Yizhong Wang, Akari Asai,
  Gabriel Ilharco, Hannaneh Hajishirzi, and Jonathan Berant.
\newblock 2021.
\newblock Multimodalqa: complex question answering over text, tables and
  images.
\newblock In {\em 9th International Conference on Learning Representations,
  {ICLR} 2021, Virtual Event, Austria, May 3-7, 2021}. OpenReview.net.

\bibitem[\protect\citename{Thorne \bgroup et al.\egroup
  }2018]{DBLP:conf/naacl/ThorneVCM18}
James Thorne, Andreas Vlachos, Christos Christodoulopoulos, and Arpit Mittal.
\newblock 2018.
\newblock {FEVER:} a large-scale dataset for fact extraction and verification.
\newblock In Marilyn~A. Walker, Heng Ji, and Amanda Stent, editors, {\em
  Proceedings of the 2018 Conference of the North American Chapter of the
  Association for Computational Linguistics: Human Language Technologies,
  {NAACL-HLT} 2018, New Orleans, Louisiana, USA, June 1-6, 2018, Volume 1 (Long
  Papers)}, pages 809--819. Association for Computational Linguistics.

\bibitem[\protect\citename{Vaicenavicius \bgroup et al.\egroup
  }2019]{DBLP:conf/aistats/VaicenaviciusWA19}
Juozas Vaicenavicius, David Widmann, Carl~R. Andersson, Fredrik Lindsten, Jacob
  Roll, and Thomas~B. Sch{\"{o}}n.
\newblock 2019.
\newblock Evaluating model calibration in classification.
\newblock In Kamalika Chaudhuri and Masashi Sugiyama, editors, {\em The 22nd
  International Conference on Artificial Intelligence and Statistics, {AISTATS}
  2019, 16-18 April 2019, Naha, Okinawa, Japan}, volume~89 of {\em Proceedings
  of Machine Learning Research}, pages 3459--3467. {PMLR}.

\bibitem[\protect\citename{Valmeekam \bgroup et al.\egroup
  }2022]{DBLP:journals/corr/abs-2206-10498}
Karthik Valmeekam, Alberto~Olmo Hernandez, Sarath Sreedharan, and Subbarao
  Kambhampati.
\newblock 2022.
\newblock Large language models still can't plan {(A} benchmark for llms on
  planning and reasoning about change).
\newblock {\em CoRR}, abs/2206.10498.

\bibitem[\protect\citename{Valmeekam \bgroup et al.\egroup
  }2023]{DBLP:journals/corr/abs-2305-15771}
Karthik Valmeekam, Matthew Marquez, Sarath Sreedharan, and Subbarao
  Kambhampati.
\newblock 2023.
\newblock On the planning abilities of large language models - {A} critical
  investigation.
\newblock {\em CoRR}, abs/2305.15771.

\bibitem[\protect\citename{Vowels \bgroup et al.\egroup
  }2023]{DBLP:journals/csur/VowelsCB23}
Matthew~J. Vowels, Necati~Cihan Camg{\"{o}}z, and Richard Bowden.
\newblock 2023.
\newblock D'ya like dags? {A} survey on structure learning and causal
  discovery.
\newblock {\em {ACM} Comput. Surv.}, 55(4):82:1--82:36.

\bibitem[\protect\citename{Wadden \bgroup et al.\egroup
  }2020]{DBLP:conf/emnlp/WaddenLLWZCH20}
David Wadden, Shanchuan Lin, Kyle Lo, Lucy~Lu Wang, Madeleine van Zuylen, Arman
  Cohan, and Hannaneh Hajishirzi.
\newblock 2020.
\newblock Fact or fiction: Verifying scientific claims.
\newblock In Bonnie Webber, Trevor Cohn, Yulan He, and Yang Liu, editors, {\em
  Proceedings of the 2020 Conference on Empirical Methods in Natural Language
  Processing, {EMNLP} 2020, Online, November 16-20, 2020}, pages 7534--7550.
  Association for Computational Linguistics.

\bibitem[\protect\citename{Wang \bgroup et al.\egroup
  }2019a]{DBLP:conf/nips/WangPNSMHLB19}
Alex Wang, Yada Pruksachatkun, Nikita Nangia, Amanpreet Singh, Julian Michael,
  Felix Hill, Omer Levy, and Samuel~R. Bowman.
\newblock 2019a.
\newblock Superglue: {A} stickier benchmark for general-purpose language
  understanding systems.
\newblock In Hanna~M. Wallach, Hugo Larochelle, Alina Beygelzimer, Florence
  d'Alch{\'{e}}{-}Buc, Emily~B. Fox, and Roman Garnett, editors, {\em Advances
  in Neural Information Processing Systems 32: Annual Conference on Neural
  Information Processing Systems 2019, NeurIPS 2019, December 8-14, 2019,
  Vancouver, BC, Canada}, pages 3261--3275.

\bibitem[\protect\citename{Wang \bgroup et al.\egroup
  }2019b]{DBLP:conf/iclr/WangSMHLB19}
Alex Wang, Amanpreet Singh, Julian Michael, Felix Hill, Omer Levy, and
  Samuel~R. Bowman.
\newblock 2019b.
\newblock {GLUE:} {A} multi-task benchmark and analysis platform for natural
  language understanding.
\newblock In {\em 7th International Conference on Learning Representations,
  {ICLR} 2019, New Orleans, LA, USA, May 6-9, 2019}. OpenReview.net.

\bibitem[\protect\citename{Wang \bgroup et al.\egroup
  }2022a]{DBLP:journals/corr/abs-2212-13465}
Dingzirui Wang, Longxu Dou, and Wanxiang Che.
\newblock 2022a.
\newblock A survey on table-and-text hybridqa: Concepts, methods, challenges
  and future directions.
\newblock {\em CoRR}, abs/2212.13465.

\bibitem[\protect\citename{Wang \bgroup et al.\egroup
  }2022b]{DBLP:journals/corr/abs-2205-11097}
Lijie Wang, Yaozong Shen, Shuyuan Peng, Shuai Zhang, Xinyan Xiao, Hao Liu,
  Hongxuan Tang, Ying Chen, Hua Wu, and Haifeng Wang.
\newblock 2022b.
\newblock A fine-grained interpretability evaluation benchmark for neural
  {NLP}.
\newblock In {\em Proceedings of the 26th Conference on Computational Natural
  Language Learning (CoNLL)}, pages 70--84, Abu Dhabi, United Arab Emirates
  (Hybrid), December. Association for Computational Linguistics.

\bibitem[\protect\citename{Wang \bgroup et al.\egroup
  }2022c]{DBLP:conf/emnlp/WangC0PWL00LLLZ22}
Xiaozhi Wang, Yulin Chen, Ning Ding, Hao Peng, Zimu Wang, Yankai Lin, Xu~Han,
  Lei Hou, Juanzi Li, Zhiyuan Liu, Peng Li, and Jie Zhou.
\newblock 2022c.
\newblock {MAVEN-ERE:} {A} unified large-scale dataset for event coreference,
  temporal, causal, and subevent relation extraction.
\newblock In Yoav Goldberg, Zornitsa Kozareva, and Yue Zhang, editors, {\em
  Proceedings of the 2022 Conference on Empirical Methods in Natural Language
  Processing, {EMNLP} 2022, Abu Dhabi, United Arab Emirates, December 7-11,
  2022}, pages 926--941. Association for Computational Linguistics.

\bibitem[\protect\citename{Wang \bgroup et al.\egroup
  }2022d]{DBLP:conf/naacl/0002WY22}
Xuezhi Wang, Haohan Wang, and Diyi Yang.
\newblock 2022d.
\newblock Measure and improve robustness in {NLP} models: {A} survey.
\newblock In Marine Carpuat, Marie{-}Catherine de~Marneffe, and Iv{\'{a}}n
  Vladimir~Meza Ru{\'{\i}}z, editors, {\em Proceedings of the 2022 Conference
  of the North American Chapter of the Association for Computational
  Linguistics: Human Language Technologies, {NAACL} 2022, Seattle, WA, United
  States, July 10-15, 2022}, pages 4569--4586. Association for Computational
  Linguistics.

\bibitem[\protect\citename{Wang \bgroup et al.\egroup }2023]{sc_chatgpt_1}
Zengzhi Wang, Qiming Xie, Zixiang Ding, Yi~Feng, and Rui Xia.
\newblock 2023.
\newblock Is chatgpt a good sentiment analyzer? {A} preliminary study.
\newblock {\em CoRR}, abs/2304.04339.

\bibitem[\protect\citename{Warstadt \bgroup et al.\egroup
  }2020]{DBLP:journals/tacl/WarstadtPLMPWB20}
Alex Warstadt, Alicia Parrish, Haokun Liu, Anhad Mohananey, Wei Peng,
  Sheng{-}Fu Wang, and Samuel~R. Bowman.
\newblock 2020.
\newblock Blimp: The benchmark of linguistic minimal pairs for english.
\newblock {\em Trans. Assoc. Comput. Linguistics}, 8:377--392.

\bibitem[\protect\citename{Waseem \bgroup et al.\egroup
  }2017]{UnderstandingAbuse}
Zeerak Waseem, Thomas Davidson, Dana Warmsley, and Ingmar Weber.
\newblock 2017.
\newblock Understanding abuse: A typology of abusive language detection
  subtasks.
\newblock In {\em Proceedings of the First Workshop on Abusive Language
  Online}, pages 78--84, Vancouver, BC, Canada, August. Association for
  Computational Linguistics.

\bibitem[\protect\citename{Wei \bgroup et al.\egroup
  }2022a]{DBLP:journals/tmlr/WeiTBRZBYBZMCHVLDF22}
Jason Wei, Yi~Tay, Rishi Bommasani, Colin Raffel, Barret Zoph, Sebastian
  Borgeaud, Dani Yogatama, Maarten Bosma, Denny Zhou, Donald Metzler, Ed~H.
  Chi, Tatsunori Hashimoto, Oriol Vinyals, Percy Liang, Jeff Dean, and William
  Fedus.
\newblock 2022a.
\newblock Emergent abilities of large language models.
\newblock {\em Trans. Mach. Learn. Res.}, 2022.

\bibitem[\protect\citename{Wei \bgroup et al.\egroup
  }2022b]{DBLP:conf/nips/Wei0SBIXCLZ22}
Jason Wei, Xuezhi Wang, Dale Schuurmans, Maarten Bosma, Brian Ichter, Fei Xia,
  Ed~H. Chi, Quoc~V. Le, and Denny Zhou.
\newblock 2022b.
\newblock Chain-of-thought prompting elicits reasoning in large language
  models.
\newblock In {\em NeurIPS}.

\bibitem[\protect\citename{Weidinger \bgroup et al.\egroup
  }2021]{DBLP:journals/corr/abs-2112-04359}
Laura Weidinger, John Mellor, Maribeth Rauh, Conor Griffin, Jonathan Uesato,
  Po{-}Sen Huang, Myra Cheng, Mia Glaese, Borja Balle, Atoosa Kasirzadeh, Zac
  Kenton, Sasha Brown, Will Hawkins, Tom Stepleton, Courtney Biles, Abeba
  Birhane, Julia Haas, Laura Rimell, Lisa~Anne Hendricks, William Isaac, Sean
  Legassick, Geoffrey Irving, and Iason Gabriel.
\newblock 2021.
\newblock Ethical and social risks of harm from language models.
\newblock {\em CoRR}, abs/2112.04359.

\bibitem[\protect\citename{Wu \bgroup et al.\egroup
  }2021]{DBLP:conf/icml/WuRLBGS21}
Yuhuai Wu, Markus~N. Rabe, Wenda Li, Jimmy Ba, Roger~B. Grosse, and Christian
  Szegedy.
\newblock 2021.
\newblock {LIME:} learning inductive bias for primitives of mathematical
  reasoning.
\newblock In Marina Meila and Tong Zhang, editors, {\em Proceedings of the 38th
  International Conference on Machine Learning, {ICML} 2021, 18-24 July 2021,
  Virtual Event}, volume 139 of {\em Proceedings of Machine Learning Research},
  pages 11251--11262. {PMLR}.

\bibitem[\protect\citename{Wulczyn \bgroup et al.\egroup }2017]{ExMachina:}
Ellery Wulczyn, Nithum Thain, and Lucas Dixon.
\newblock 2017.
\newblock Ex machina: Personal attacks seen at scale.
\newblock In {\em Proceedings of the 26th International Conference on World
  Wide Web}, WWW '17, page 1391–1399, Republic and Canton of Geneva, CHE.
  International World Wide Web Conferences Steering Committee.

\bibitem[\protect\citename{Xie \bgroup et al.\egroup
  }2022]{DBLP:conf/emnlp/XieW0ZSYWZYWZWL22}
Tianbao Xie, Chen~Henry Wu, Peng Shi, Ruiqi Zhong, Torsten Scholak, Michihiro
  Yasunaga, Chien{-}Sheng Wu, Ming Zhong, Pengcheng Yin, Sida~I. Wang, Victor
  Zhong, Bailin Wang, Chengzu Li, Connor Boyle, Ansong Ni, Ziyu Yao, Dragomir
  Radev, Caiming Xiong, Lingpeng Kong, Rui Zhang, Noah~A. Smith, Luke
  Zettlemoyer, and Tao Yu.
\newblock 2022.
\newblock Unifiedskg: Unifying and multi-tasking structured knowledge grounding
  with text-to-text language models.
\newblock In Yoav Goldberg, Zornitsa Kozareva, and Yue Zhang, editors, {\em
  Proceedings of the 2022 Conference on Empirical Methods in Natural Language
  Processing, {EMNLP} 2022, Abu Dhabi, United Arab Emirates, December 7-11,
  2022}, pages 602--631. Association for Computational Linguistics.

\bibitem[\protect\citename{Xu \bgroup et al.\egroup }2021]{BotAdversarial}
Jing Xu, Da~Ju, Margaret Li, Y-Lan Boureau, Jason Weston, and Emily Dinan.
\newblock 2021.
\newblock Bot-adversarial dialogue for safe conversational agents.
\newblock In {\em Proceedings of the 2021 Conference of the North American
  Chapter of the Association for Computational Linguistics: Human Language
  Technologies}, pages 2950--2968, Online, June. Association for Computational
  Linguistics.

\bibitem[\protect\citename{Yang \bgroup et al.\egroup
  }2015]{DBLP:conf/emnlp/YangYM15}
Yi~Yang, Wen{-}tau Yih, and Christopher Meek.
\newblock 2015.
\newblock Wikiqa: {A} challenge dataset for open-domain question answering.
\newblock In Llu{\'{\i}}s M{\`{a}}rquez, Chris Callison{-}Burch, Jian Su,
  Daniele Pighin, and Yuval Marton, editors, {\em Proceedings of the 2015
  Conference on Empirical Methods in Natural Language Processing, {EMNLP} 2015,
  Lisbon, Portugal, September 17-21, 2015}, pages 2013--2018. The Association
  for Computational Linguistics.

\bibitem[\protect\citename{Yang \bgroup et al.\egroup
  }2018]{DBLP:conf/emnlp/Yang0ZBCSM18}
Zhilin Yang, Peng Qi, Saizheng Zhang, Yoshua Bengio, William~W. Cohen, Ruslan
  Salakhutdinov, and Christopher~D. Manning.
\newblock 2018.
\newblock Hotpotqa: {A} dataset for diverse, explainable multi-hop question
  answering.
\newblock In Ellen Riloff, David Chiang, Julia Hockenmaier, and Jun'ichi
  Tsujii, editors, {\em Proceedings of the 2018 Conference on Empirical Methods
  in Natural Language Processing, Brussels, Belgium, October 31 - November 4,
  2018}, pages 2369--2380. Association for Computational Linguistics.

\bibitem[\protect\citename{Yang \bgroup et al.\egroup
  }2022]{DBLP:journals/corr/abs-2212-10923}
Zonglin Yang, Li~Dong, Xinya Du, Hao Cheng, Erik Cambria, Xiaodong Liu,
  Jianfeng Gao, and Furu Wei.
\newblock 2022.
\newblock Language models as inductive reasoners.
\newblock {\em CoRR}, abs/2212.10923.

\bibitem[\protect\citename{Yasunaga \bgroup et al.\egroup
  }2022]{DBLP:conf/acl/YasunagaLL22}
Michihiro Yasunaga, Jure Leskovec, and Percy Liang.
\newblock 2022.
\newblock Linkbert: Pretraining language models with document links.
\newblock In Smaranda Muresan, Preslav Nakov, and Aline Villavicencio, editors,
  {\em Proceedings of the 60th Annual Meeting of the Association for
  Computational Linguistics (Volume 1: Long Papers), {ACL} 2022, Dublin,
  Ireland, May 22-27, 2022}, pages 8003--8016. Association for Computational
  Linguistics.

\bibitem[\protect\citename{Ye and Durrett}2022]{DBLP:conf/nips/YeD22}
Xi~Ye and Greg Durrett.
\newblock 2022.
\newblock The unreliability of explanations in few-shot prompting for textual
  reasoning.
\newblock In {\em NeurIPS}.

\bibitem[\protect\citename{Yin \bgroup et al.\egroup
  }2023]{DBLP:journals/corr/abs-2305-18153}
Zhangyue Yin, Qiushi Sun, Qipeng Guo, Jiawen Wu, Xipeng Qiu, and Xuanjing
  Huang.
\newblock 2023.
\newblock Do large language models know what they don't know?
\newblock {\em CoRR}, abs/2305.18153.

\bibitem[\protect\citename{Yu \bgroup et al.\egroup }2023]{yu2023natural}
Fei Yu, Hongbo Zhang, Prayag Tiwari, and Benyou Wang.
\newblock 2023.
\newblock Natural language reasoning, a survey.

\bibitem[\protect\citename{Zampieri \bgroup et al.\egroup
  }2019]{Predictingthetypeand}
Marcos Zampieri, Shervin Malmasi, Preslav Nakov, Sara Rosenthal, Noura Farra,
  and Ritesh Kumar.
\newblock 2019.
\newblock Predicting the type and target of offensive posts in social media.
\newblock In {\em Proceedings of the 2019 Conference of the North {A}merican
  Chapter of the Association for Computational Linguistics: Human Language
  Technologies, Volume 1 (Long and Short Papers)}, pages 1415--1420,
  Minneapolis, Minnesota, June. Association for Computational Linguistics.

\bibitem[\protect\citename{Zelikman \bgroup et al.\egroup
  }2022]{Zelikman2022STaRBR}
E.~Zelikman, Yuhuai Wu, and Noah~D. Goodman.
\newblock 2022.
\newblock Star: Bootstrapping reasoning with reasoning.
\newblock {\em ArXiv}, abs/2203.14465.

\bibitem[\protect\citename{Zellers \bgroup et al.\egroup
  }2019]{DBLP:conf/acl/ZellersHBFC19}
Rowan Zellers, Ari Holtzman, Yonatan Bisk, Ali Farhadi, and Yejin Choi.
\newblock 2019.
\newblock Hellaswag: Can a machine really finish your sentence?
\newblock In Anna Korhonen, David~R. Traum, and Llu{\'{\i}}s M{\`{a}}rquez,
  editors, {\em Proceedings of the 57th Conference of the Association for
  Computational Linguistics, {ACL} 2019, Florence, Italy, July 28- August 2,
  2019, Volume 1: Long Papers}, pages 4791--4800. Association for Computational
  Linguistics.

\bibitem[\protect\citename{Zeng and Li}2022]{iron}
Qingcheng Zeng and An{-}Ran Li.
\newblock 2022.
\newblock A survey in automatic irony processing: Linguistic, cognitive, and
  multi-x perspectives.
\newblock In Nicoletta Calzolari, Chu{-}Ren Huang, Hansaem Kim, James
  Pustejovsky, Leo Wanner, Key{-}Sun Choi, Pum{-}Mo Ryu, Hsin{-}Hsi Chen, Lucia
  Donatelli, Heng Ji, Sadao Kurohashi, Patrizia Paggio, Nianwen Xue, Seokhwan
  Kim, Younggyun Hahm, Zhong He, Tony~Kyungil Lee, Enrico Santus, Francis Bond,
  and Seung{-}Hoon Na, editors, {\em Proceedings of the 29th International
  Conference on Computational Linguistics, {COLING} 2022, Gyeongju, Republic of
  Korea, October 12-17, 2022}, pages 824--836. International Committee on
  Computational Linguistics.

\bibitem[\protect\citename{Zhang \bgroup et al.\egroup }2015]{dataset_yelp5}
Xiang Zhang, Junbo~Jake Zhao, and Yann LeCun.
\newblock 2015.
\newblock Character-level convolutional networks for text classification.
\newblock In Corinna Cortes, Neil~D. Lawrence, Daniel~D. Lee, Masashi Sugiyama,
  and Roman Garnett, editors, {\em Advances in Neural Information Processing
  Systems 28: Annual Conference on Neural Information Processing Systems 2015,
  December 7-12, 2015, Montreal, Quebec, Canada}, pages 649--657.

\bibitem[\protect\citename{Zhang \bgroup et al.\egroup }2022]{absa_survey}
Wenxuan Zhang, Xin Li, Yang Deng, Lidong Bing, and Wai Lam.
\newblock 2022.
\newblock A survey on aspect-based sentiment analysis: Tasks, methods, and
  challenges.
\newblock {\em CoRR}, abs/2203.01054.

\bibitem[\protect\citename{Zhang \bgroup et al.\egroup }2023]{sc_llm_1}
Wenxuan Zhang, Yue Deng, Bing Liu, Sinno~Jialin Pan, and Lidong Bing.
\newblock 2023.
\newblock Sentiment analysis in the era of large language models: {A} reality
  check.
\newblock {\em CoRR}, abs/2305.15005.

\bibitem[\protect\citename{Zhao and Vydiswaran}2021]{DBLP:conf/aaai/ZhaoV21}
Xinyan Zhao and V.~G.~Vinod Vydiswaran.
\newblock 2021.
\newblock Lirex: Augmenting language inference with relevant explanations.
\newblock In {\em Thirty-Fifth {AAAI} Conference on Artificial Intelligence,
  {AAAI} 2021, Thirty-Third Conference on Innovative Applications of Artificial
  Intelligence, {IAAI} 2021, The Eleventh Symposium on Educational Advances in
  Artificial Intelligence, {EAAI} 2021, Virtual Event, February 2-9, 2021},
  pages 14532--14539. {AAAI} Press.

\bibitem[\protect\citename{Zhao \bgroup et al.\egroup
  }2023]{DBLP:journals/corr/abs-2303-18223}
Wayne~Xin Zhao, Kun Zhou, Junyi Li, Tianyi Tang, Xiaolei Wang, Yupeng Hou,
  Yingqian Min, Beichen Zhang, Junjie Zhang, Zican Dong, Yifan Du, Chen Yang,
  Yushuo Chen, Zhipeng Chen, Jinhao Jiang, Ruiyang Ren, Yifan Li, Xinyu Tang,
  Zikang Liu, Peiyu Liu, Jian{-}Yun Nie, and Ji{-}Rong Wen.
\newblock 2023.
\newblock A survey of large language models.
\newblock {\em CoRR}, abs/2303.18223.

\bibitem[\protect\citename{Zheng \bgroup et al.\egroup }2023]{zheng2023does}
Shen Zheng, Jie Huang, and Kevin Chen-Chuan Chang.
\newblock 2023.
\newblock Why does chatgpt fall short in providing truthful answers?

\bibitem[\protect\citename{Zhong \bgroup et al.\egroup
  }2017]{DBLP:journals/corr/abs-1709-00103}
Victor Zhong, Caiming Xiong, and Richard Socher.
\newblock 2017.
\newblock Seq2sql: Generating structured queries from natural language using
  reinforcement learning.
\newblock {\em CoRR}, abs/1709.00103.

\bibitem[\protect\citename{Zhong \bgroup et al.\egroup
  }2023]{DBLP:journals/corr/abs-2304-06364}
Wanjun Zhong, Ruixiang Cui, Yiduo Guo, Yaobo Liang, Shuai Lu, Yanlin Wang, Amin
  Saied, Weizhu Chen, and Nan Duan.
\newblock 2023.
\newblock Agieval: {A} human-centric benchmark for evaluating foundation
  models.
\newblock {\em CoRR}, abs/2304.06364.

\bibitem[\protect\citename{Zhou \bgroup et al.\egroup }2018]{ecm}
Hao Zhou, Minlie Huang, Tianyang Zhang, Xiaoyan Zhu, and Bing Liu.
\newblock 2018.
\newblock Emotional chatting machine: Emotional conversation generation with
  internal and external memory.
\newblock In Sheila~A. McIlraith and Kilian~Q. Weinberger, editors, {\em
  Proceedings of the Thirty-Second {AAAI} Conference on Artificial
  Intelligence, (AAAI-18), the 30th innovative Applications of Artificial
  Intelligence (IAAI-18), and the 8th {AAAI} Symposium on Educational Advances
  in Artificial Intelligence (EAAI-18), New Orleans, Louisiana, USA, February
  2-7, 2018}, pages 730--739. {AAAI} Press.

\bibitem[\protect\citename{Zhuo \bgroup et al.\egroup }2023]{zhuo2023red}
Terry~Yue Zhuo, Yujin Huang, Chunyang Chen, and Zhenchang Xing.
\newblock 2023.
\newblock Red teaming chatgpt via jailbreaking: Bias, robustness, reliability
  and toxicity.

\bibitem[\protect\citename{Ziems \bgroup et al.\egroup
  }2022]{themoralintegrity}
Caleb Ziems, Jane Yu, Yi-Chia Wang, Alon Halevy, and Diyi Yang.
\newblock 2022.
\newblock The moral integrity corpus: A benchmark for ethical dialogue systems.
\newblock In {\em Proceedings of the 60th Annual Meeting of the Association for
  Computational Linguistics (Volume 1: Long Papers)}, pages 3755--3773, Dublin,
  Ireland, May. Association for Computational Linguistics.

\end{thebibliography}

\end{document}